%% file: main.tex
\documentclass{article}

\usepackage{microtype}
\usepackage{graphicx}
\usepackage{subfigure}
\usepackage{booktabs} 

\usepackage{comment}
\usepackage{amsmath,amssymb}
\usepackage{color}
\usepackage{epsfig}
\usepackage{booktabs}
\usepackage{multirow}
\usepackage{array}
\usepackage{adjustbox}
\usepackage{setspace}
\usepackage{wrapfig}
\usepackage{arydshln}
\usepackage{autobreak}
\usepackage{ragged2e}
\usepackage{scalerel}
\usepackage{amsthm}
\usepackage[table]{xcolor}

\usepackage{hyperref}

\newcommand{\vect}[1]{\boldsymbol{\mathbf{#1}}}

\usepackage[accepted]{icml2023}

\usepackage{amsmath}
\usepackage{amssymb}
\usepackage{mathtools}
\usepackage{amsthm}

\usepackage[capitalize,noabbrev]{cleveref}

\theoremstyle{plain}

\theoremstyle{definition}

\theoremstyle{remark}

\usepackage[textsize=tiny]{todonotes}

\icmltitlerunning{Training Full Spike Neural Networks with Auxiliary Accumulation Pathway}

\begin{document}

\twocolumn[
\icmltitle{Training Full Spike Neural Networks via Auxiliary Accumulation Pathway}



\icmlsetsymbol{equal}{*}

\begin{icmlauthorlist}
\icmlauthor{Guangyao Chen}{pku,pcl}
\icmlauthor{Peixi Peng}{pku,pcl}
\icmlauthor{Guoqi LI}{auto}
\icmlauthor{Yonghong Tian}{pku,pcl}
\end{icmlauthorlist}

\icmlaffiliation{pku}{Department of Computer Science and Technology, Peking University}
\icmlaffiliation{pcl}{Peng Cheng Laborotory}
\icmlaffiliation{auto}{Institute of Automation, Chinese Academy of Sciences}

\icmlcorrespondingauthor{Peixi Peng}{pxpeng@pku.edu.cn}
\icmlcorrespondingauthor{Yonghong Tian}{yhtian@pku.edu.cn}

\icmlkeywords{Machine Learning, ICML}

\vskip 0.3in
]



\printAffiliationsAndNotice{\icmlEqualContribution} 

\begin{abstract}
Due to the binary spike signals making converting the traditional high-power multiply-accumulation (MAC) into a low-power accumulation (AC) available,  the brain-inspired Spiking Neural Networks (SNNs) are gaining more and more attention. 
 However, the binary spike propagation of the Full-Spike Neural Networks (FSNN) with limited time steps is prone to significant information loss.
To improve performance, several state-of-the-art SNN models trained from scratch inevitably bring many non-spike operations. The non-spike operations cause additional computational consumption and may not be deployed on some neuromorphic hardware where only spike operation is allowed.
To train a large-scale FSNN with high performance, this paper proposes a novel Dual-Stream Training (DST) method which adds a detachable Auxiliary Accumulation Pathway (AAP) to the full spiking residual networks.
 The accumulation in  AAP  could compensate for the information loss during the forward and backward of full spike propagation, and facilitate the training of the FSNN. In the test phase, the AAP could be removed and only the FSNN is remained. This not only keeps the lower energy consumption but also makes our model easy to deploy. Moreover, for some cases where the non-spike operations are available,  the APP could also be retained in  test inference and improve feature discrimination by introducing a little non-spike consumption.
 Extensive experiments on ImageNet, DVS Gesture, and CIFAR10-DVS datasets demonstrate the effectiveness of DST. 
\end{abstract}

\input{Main/Tex/01_introduction}
\input{Main/Tex/02_related_works}
\input{Main/Tex/03_preliminaries}

\input{Main/Tex/04_methodology}

\input{Main/Tex/05_experiments}

\input{Main/Tex/06_conclusion}

\bibliography{ref}
\bibliographystyle{icml2023}

\newpage
\appendix
\onecolumn

\input{Main/Tex/07_appendix}


\end{document}

%% file: Main/Tex/01_introduction.tex
\section{Introduction}
\label{introduction}

In the past few years, Artificial Neural Networks (ANNs) have achieved great success in many tasks  \cite{krizhevsky2012imagenet,simonyan2015deep,szegedy2015going,girshick2014rich,liu2016ssd,redmon2016you,chenadversarial,chen2021amplitude,chen2020learning,ma2022picking,chen2018saliency}.
However, with ANNs getting deeper and larger, computational and power consumption are growing rapidly. 
Hence, Spiking Neural Networks (SNNs), inspired by biological neurons, have recently received surging attention and are regarded as a potential competitor of ANNs due to their high biological plausibility, event-driven property, and low power consumption~\cite{roy2019towards} on neuromorphic hardware. 


To obtain an effective SNN, several works \cite{KIM2018373, 10.1007/978-3-030-36718-3_15, 10.3389/fnins.2021.629000, hu2020spiking, sengupta2019going,Han_2020_CVPR,lee2020enabling, zheng2020going, samadzadeh2021convolutional,rathi2020dietsnn, rathi2020enabling} are proposed to convert the trained ANN to SNN by replacing the raw activation layers (such as ReLU) with spiking neurons. Although this type of method could achieve state-of-the-art accuracy on many image classification datasets, they often require a large number of time steps which causes high computational consumption and also limit the application of SNN, and the direct promotion of exploring the characteristics of SNN is limited. Hence, a series of methods are proposed to train SNN from scratch. Based on the surrogate gradient backpropagation method \cite{TAVANAEI201947}, several recent works train deep SNN by improving the Batchnorm \cite{zheng2020going} or residual connection structure \cite{fang2021deep,zhou2023spikformer,xiaoonline,deng2021temporal}, and narrow the gap between SNN and ANN effectively.
 To obtain high performance, several SOTA SNN models~\cite{fang2021deep,zheng2020going,zhou2023spikformer} inevitably bring many non-spike operations with ADD residual connections. Although effective, these methods may suffer two main drawbacks: 
 Firstly, the energy efficiency advantage of SNNs mainly comes from the fact that the binary spike signals make converting the traditional high-power multiply-accumulation (MAC) into a low-power accumulation (AC) available, the non-spike operations don't fit this characteristic and will bring high computation consumption. Secondly, several neuromorphic hardware only supports spiking operation, and these models cannot be deployed directly~\cite{horowitz20141}. 

\begin{figure*}[!tb]
\centering
\includegraphics[width=\linewidth]{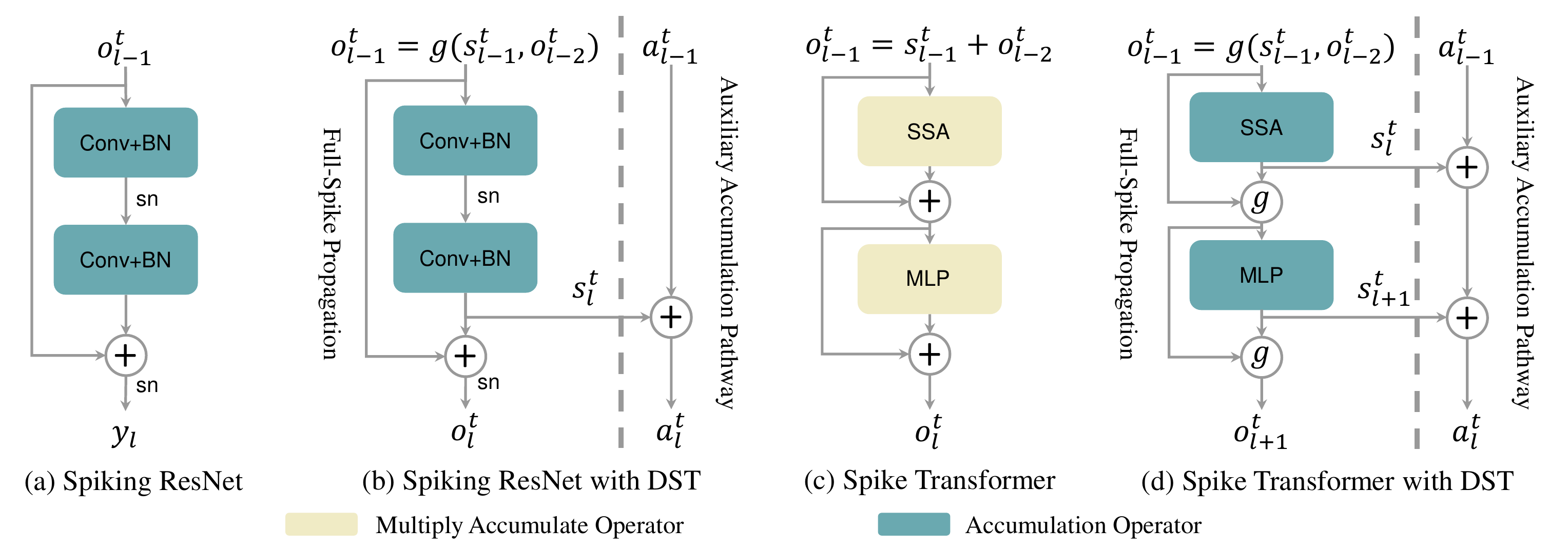}
\vspace{-0.8cm}
\caption{Residual blocks and Dual-stream blocks in Spiking ResNet~\cite{fang2021deep} and Spike Transformer~\cite{zhou2023spikformer}. (a) the Spiking ResNet by replacing ReLU
activation layers with spiking neurons. (b) The Spiking ResNet with Dual-stream training. (c) The input of the Spike Transformer with Spiking Self-Attention (SSA) is non-spike data due to residual addition between blocks, which brings additional multiply accumulative computation to SNN. (d) Spike Transformer with a dual-stream structure is divided into a spike propagation pathway and a plug-and-play auxiliary accumulation  pathway (AAP), which could compensate for full spike propagation. Note AAP could be either retained or removed in model inference.}
\label{fig:blocks}

\end{figure*}

Hence, it is necessary to develop a full-spike neural network (FSNN) that only contains spike operations. However, the binary spike propagation with limited time steps is prone to significant information loss, and limits the performance of FSNN. 
For example, the Spiking ResNet in Figure~\ref{fig:blocks} is prone to vanishing gradient problems in deep networks~\cite{fang2021deep}.
To compensate for the loss of information forward and backward from full spike propagation, we propose  a novel Dual-stream Training (DST) method, where the whole network contains a \textit{full spike propagation} stream and a \textit{auxiliary spike accumulation} stream. The former includes full-spike inference of FSNN, while the latter is  a plug-and-play Auxiliary Accumulation Pathway (AAP) to the FSNN.
In training, the AAP is able to compensate for the loss of information forward and backward from full spike propagation by spike accumulation, which could help alleviate the vanishing gradient problem of the Spiking ResNet and improve the performance of the FSNN. 
Although the accumulation in APP causes non-spike operation, the AAP could be removed and only the FSNN remains in the test phase. In other words, our model could act as an FSNN inference in practical application. 
This not only keeps the lower energy consumption but also makes our model easy to deploy in the neuromorphic hardware. 
Moreover, for some cases where the non-spike operations are available, the AAP could also be retained in  test inference and further improve feature discrimination by introducing  a small number of non-spike MAC operations. It is notable that  FSNN+AAP only brings a linear rise in the non-spike AC computation with the model depth, resulting in less computational consumption. 

We evaluate DSNN on both the static ImageNet dataset~\cite{deng2009imagenet} and the neuromorphic DVS Gesture dataset~\cite{amir2017low}, CIFAR10-DVS dataset~\cite{cifar10_dvs}, CIFAR100~ \cite{krizhevsky2009learning}. 
The experiment results are consistent with our analysis, indicating that the DST could improve the previous ResNet-based and Transformer-based FSNNs to higher performance by simply increasing the network’s depth, and keeping efficient computation consumption simultaneously.

%% file: Main/Tex/02_related_works.tex
\section{Related Work}

\subsection{Spiking Neural Networks}
To train deep SNNs, ANN to SNN conversion (ANN2SNN)~\cite{hunsberger2015spiking, cao2015spiking, Bodo2017Conversion, sengupta2019going, Han_2020_CVPR, han2020deep, deng2021optimal, stockl2021optimized, pmlr-v139-li21d} and backpropagation with surrogate gradient ~\cite{neftci2019surrogate} are the two mainstream methods. 
For ANN2SNN, an ANN with ReLU activation is first trained, then converts the ANN to an SNN by replacing ReLU with spiking neurons and adding scaling operations like weight normalization and threshold balancing.
Recently, several ANN2SNN methods~\cite{Han_2020_CVPR, han2020deep, deng2021optimal, pmlr-v139-li21d} have achieved near loss-less accuracy for VGG-16 and ResNet.
However, the converted SNN needs longer time steps to rival the original ANN, and increases the SNN's computational consumption ~\cite{Bodo2017Conversion}. 
One of the backpropagation methods~\cite{kim2020unifying} computes the gradients of the timings of existing spikes with respect to the membrane potential at the spike timing \cite{comsa2020temporal, mostafa2017supervised, kheradpisheh2020temporal, zhou2019temporal, zhang2020temporal}. 
Another kind of backpropagation method gets the gradient by unfolding the network over the simulation time-steps~\cite{lee2016training, huh2017gradient, wu2018STBP, shrestha2018slayer, lee2020enabling, neftci2019surrogate}. 
As the gradient with respect to the threshold-triggered firing is non-differentiable, the surrogate gradient is often used. 

\subsection{Spiking Residual Networks}
For ANN2SNN with ResNet, several methods made specific normalizations for conversion. 
The residual structure in ANN2SNN with scaled shortcuts is applied in SNN to match the activations of the original ResNet \cite{hu2018spiking}. 
Previous ANN2SNN methods noticed the distinction between plain feedforward ANNs and residual ANNs, and made specific normalizations for conversion. 
Then, Spike-Norm~\cite{sengupta2019going} is proposed to balance the threshold of the Spiking Neural Model and verified their method by converting ResNet to SNNs. 
Moreover, existing backpropagation-based methods use nearly the same structure as ResNet.
Several custom surrogate methods~\cite{sengupta2019going} are evaluated on shallow ResNets.
Threshold-dependent batch normalization (td-BN) is proposed to replace naive batch normalization (BN) \cite{ioffe2015batch} and successfully trained Spiking ResNet-34/50 directly with surrogate gradient by adding td-BN in shortcuts.
SEW ResNet is the first method to increase the SNNs to more than 100 layers, but its use of additive operations in residual connections leads to a none-spike structure in the deep layer of the network bringing higher computational consumption.
\cite{deng2021temporal} introduces the temporal efficient training approach to compensate for the loss of momentum in the gradient descent with SG so that the training process can converge into flatter minima with better generalizability.
\cite{xiaoonline} proposes online training through time for SNNs, which enables forward-in-time learning by tracking presynaptic activities and leveraging instantaneous loss and gradients.
Recently, \cite{zhou2023spikformer} considers leveraging both self-attention capability and biological properties of SNNs and proposes the Spiking Transformer (Spikformer).

%% file: Main/Tex/03_preliminaries.tex
\section{Preliminaries}
\label{sec:pre}

\subsection{Spiking Neuron Model}
The spiking neuron is the fundamental computing unit of SNNs. The dynamics of all kinds of spiking neurons can be described as follow:
\begin{align}
	H_t &= SN(V_{t-1}, X_t), \label{neural dynamics}\\
	S_t &= \Theta(H_t - V_{th}), \label{neural spiking}\\
	V_t &= H_t~\left( 1 - S_t \right) + V_{reset}~S_t, \label{neural reset}
\end{align}
where $X_t$ is the input at time-step $t$, $H_t$ and $V_t$ denote the membrane potential after neuronal dynamics and after the trigger of a spike at time-step $t$, respectively.
$V_{th}$ is the firing threshold,  $\Theta(x)$ is the Heaviside step function and is defined by $\Theta(x) = 1$ for $x \ge 0$ and $\Theta(x) = 0$ for $x < 0$. $S_t$ is the output spike at time-step $t$, which equals 1 if there is a spike and 0 otherwise. $V_{reset}$ denotes the reset potential. The function $SN(\cdot)$ in Eq.~\eqref{neural dynamics} describes the neuronal dynamics and takes different forms for different spiking neuron models, which include the Integrate-and-Fire (IF) model (Eq.~\eqref{IFneuron}) and Leaky Integrate-and-Fire (LIF) model (Eq.~\eqref{LIFneuron}):
\begin{align}
	H_t &= V_{t-1} + X_t, \label{IFneuron}\\
	H_t &= V_{t-1} + \frac{1}{\tau}(X_t - (V_{t-1} - V_{reset})),   \label{LIFneuron} 
\end{align}
where $\tau$ represents the membrane time constant. Eq.~\eqref{neural spiking} and Eq.~\eqref{neural reset} describe the spike generation and resetting processes, which are the same for all kinds of spiking neuron models. 

\subsection{Computational Consumption}
With the sparsity of firing and the short simulation period, SNN can achieve the calculation with about the same number of synaptic operations (SyOPs) \cite{Bodo2017Conversion} rather than FLOPs, 
The number of synaptic operations per layer of the network can be easily estimated for an ANN from the architecture of the convolutional and linear layers. 
For the ANN, a multiply-accumulate (MAC) computation takes place per synaptic operation. 
On the other hand, specialized SNN hardware would perform an accumulated computation (AC) per synaptic operation only upon the receipt of an incoming spike.
Hence, the total number of AC operations occurring in the SNN would be represented by the dot-product of the average cumulative neural spike count for a particular layer and the corresponding number of synaptic operations.
With the deepening of SNN and ANN, the relative energy ratio gradually approaches a fixed value, which could be calculated as follows:
\begin{align}
\frac{E(\mathrm{SNN})}{E(\mathrm{ANN})} \thickapprox T \cdot fr \cdot \frac{E_{ac}}{E_{mac}},
\label{eq:raw_compute}
\end{align}
where $T$ and $fr$ represent the simulation time and the average firing rate.

Eq.~\eqref{eq:raw_compute} assumes the SNN only contains $\{0,1\}$ spike AC operators. 
However, several SOTA SNNs employ many non-spike MAC and their computation consumption could not be estimated by Eq.~\eqref{eq:raw_compute} accurately.
To estimate the computation consumption of different SNNs more accurately, we calculate the number of computations required in terms of AC and MAC synaptic operations, respectively.
The main energy consumption of non-spike signals comes from the MAC between neurons. 
The contribution from one neuron to another requires a MAC for each timestep, multiplying each non-spike activation with the respective weight before adding it to the internal sum.
In contrast, a transmitted spike requires only an accumulation at the target neuron, adding weight to the potential, and where spikes may be quite sparse.
Therefore, for any SNN network $\mathcal{F}$, the theoretical computational consumption can be determined by the number of AC and MAC operations ($O_{ac}, O_{mac}$):
\begin{equation}
 \begin{aligned}
    \mathrm{E}(\mathcal{F}) &= T \cdot (fr \cdot E_{ac} \cdot O_{ac} + E_{mac} \cdot O_{mac}).
 \end{aligned}
 \end{equation}
Moreover, we developed and open-sourced a tool to calculate the dynamic consumption of SNNs as \textbf{syops-counter}\footnote{\href{https://github.com/iCGY96/syops-counter}{github.com/iCGY96/syops-counter}}, which can compute the theoretical amount of AC and MAC operations.

\subsection{Spiking Residual Blocks}
\label{sec:spiking resnet}
There are two main types of residual blocks for existing SNNs. The one replaces ReLU activation layers with spiking neurons, which constructs FSNN but is prone to vanishing gradient problems. The second uses the same addition operation as the ANN, but leads to additional computation consumption due to the appearance of non-spike signals.

\paragraph{Vanishing Gradient Problems of Spiking ResNet.}
Consider a Spiking ResNet with $k$ sequential blocks to transmit $s_{l}^{t}$, and the identity mapping condition is met by using the IF neurons for residual connection with $0 < V_{th} \leq 1$, then we have $s_{l}^{t} = s_{l+1}^{t} = ... = s_{l+k-1}^{t} = o_{l+k-1}^{t}$. 
The gradient of the output of the $(l+k-1)$-th residual block with respect to the input of the $l$-th residual block could be calculated layer by layer:
\begin{equation}
\label{eqn:energy}
 \begin{aligned}
 \frac{\partial o_{l+k-1}^t}{\partial s_{l}^{t}} &= \prod_{i=0}^{k-1} \frac{\partial o_{l+i}^{t}}{\partial s_{l+i}^{t}}  = \prod_{i=0}^{k-1}\Theta'(s_{l+i}^{t} - V_{th}) \\
    & \to 
	\begin{cases}
		0, \text{if}~0 < \Theta'(s_{l}^{t} - V_{th}) < 1 \\
		1, \text{if}~\Theta'(s_{l}^{t} - V_{th}) = 1 \\
	\end{cases},
 \end{aligned}
\end{equation}
where $\Theta(x)$ is the Heaviside step function and $\Theta'(x)$ is defined by the surrogate gradient. The second equality hold as $o_{l+i}^{t} = {\rm SN}(s_{l+i}^{t})$. 
In view of the fact that $s_{l}^{t}$ could only take $0$ or $1$ with identity mapping,  $\Theta'(s_{l}^{t} - V_{th}) =1$ is not satisfied for commonly used surrogate functions mentioned in \cite{neftci2019surrogate}. 
When using the common $Sigmoid$ \cite{neftci2019surrogate} function as the Heaviside step function, the gradient vanishing problem would be prone be happen.

\paragraph{Spiking Residual Blocks with ADD.}
As illustrated in Figure~\ref{fig:blocks}(c), the residual block for SNN can be formulated with an ADD function~\cite{fang2021deep,zhou2023spikformer}, which can implement identity mapping and overcome the vanishing and exploding gradient problems.
\begin{align}
	o_l^t = \mathrm{SN}(f_l(o_{l-1}^t)) + o_{l-1}^t = s_l^t + o_{l-1}^t,
\end{align}
where $s_l^t$ denotes the residual mapping learned as $s_l^t = \mathrm{SN}(f_l(o_{l-1}^t))$.
While this design brings performance improvements, it inevitably brings in non-spike data and thus MAC operations. 
In particular, for the ADD function, if both $s_l^t$ and $o_{l-1}^t$ are spike signals, its output $o_l^t$ will be a non-spike signal whose value range is $\{0,1,2\}$. 
As the depth of the network increases, the range of signals transmitted to the next layer of the network will also expand. 
Convolution requires much more computational overhead of multiplication and addition when dealing with these non-spiking signals as shown in Figure~\ref{fig:blocks}(b). 
In this case, the network will incur additional high computational consumption.

%% file: Main/Tex/04_methodology.tex
\section{Dual-stream Training}
\label{dsnn}

\subsection{Dual-stream SNN}


\paragraph{Basic Block.}
Instead of the residual Block containing only one path with respect to the input spike $x$, we initialize the input to two consistent paths $s_0^t = a_0^t = x$, where $s_l^t$ represents the spike signal propagated between blocks, and $a_l^t$ represents the spike accumulation carried out on the output of each block.
As illustrated in Figure~\ref{fig:blocks}(b) and (d), the Dual-stream Block can be formulated as:
\begin{align}
    o_l^t &= 
    \begin{cases}
        \mathrm{SN}(f_l(o_{l-1}^t) + o_{l-1}^t) = SN(s_l^t + o_{l-1}^t) \\
        g(\mathrm{SN}(f_l(o_{l-1}^t)), o_{l-1}^t) = g(s_l^t, o_{l-1}^t) \\
    \end{cases} \label{eqn:spike route} \\
    a_l^t &=  s_l^t + a_{l-1}^t, \label{eqn:accumulation route} 
\end{align}
where $g$ represents an element-wise function with two spikes tensors as inputs.
Note that here we restrict $g$ to be only the corresponding \textit{logical operation function} as shown in Table~\ref{tab:g}, so as to ensure that the input and output of $g$ function are spike trains.
Here, Eq.\eqref{eqn:spike route} is the full-spike propagation pathway, which could be either of two types of spiking ResNet.
Eq.\eqref{eqn:accumulation route} denotes the plug-and-play Auxiliary Accumulation Pathway (AAP). 
During the inference phase, the auxiliary accumulation could be removed as needed.

\begin{figure}[!tb]
    \centering
    \setlength{\abovecaptionskip}{0.cm}
    \includegraphics[width=\linewidth]{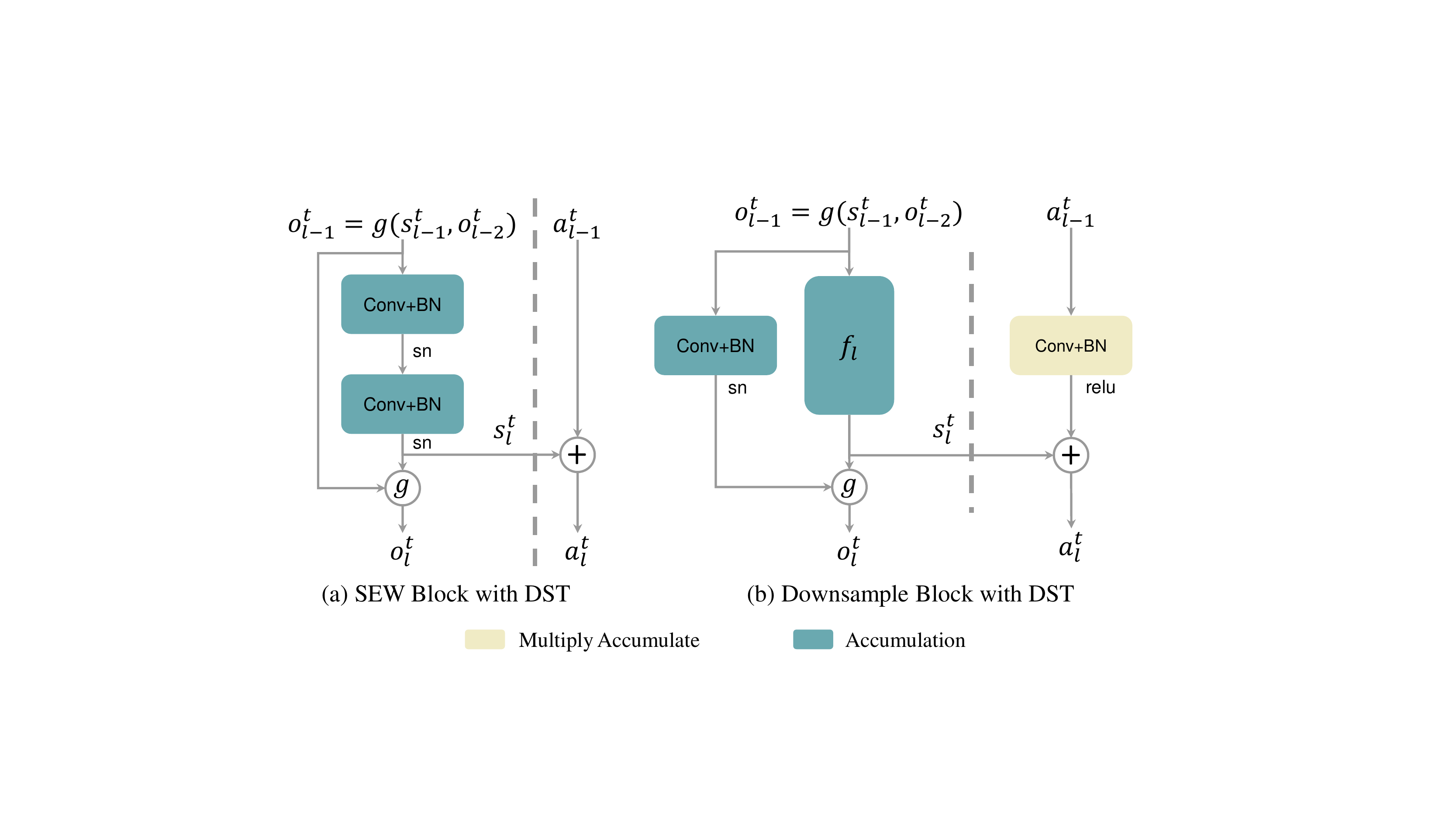}
    \vspace{-0.5cm}
    \caption{The SEW Block~\cite{fang2021deep} and its Downsample blocks with a dual-stream structure.}
    \label{fig:downsample}
    \vspace{-0.5cm}
\end{figure}

\paragraph{Downsampling Block.}
Remarkably, when the input and output of one block have different dimensions, the shortcut is set as convolutional layers with stride $>1$, rather than the identity connection, to perform downsampling. 
The Spiking ResNet utilize \{Conv-BN\} without ReLU in the shortcut. SEW ResNet \cite{fang2021deep} adds an SN in shortcut as shown in Figure~\ref{fig:downsample}(b).
Figure~\ref{fig:downsample}(b) shows the downsampling of auxiliary accumulation,  that the overhead of multiply accumulation in Dual-stream Blocks mainly comes from the downsampling of the spike accumulation signal. 
Fortunately, the number of downsampling in a network is always fixed, so the MAC-based downsampling operation of the spike accumulation pathway will not increase with the increase of network depth.
Therefore, the increased computational overhead of DSNN with the increase of network depth mainly comes from the accumulation calculation.

\noindent \textbf{Training.}
For backpropagation, the gradient could be backpropagated to these spiking neurons through the auxiliary accumulation to prevent the vanishing gradient caused by deeper layers. 
Therefore, in the training phase, the outputs of both auxiliary accumulation and spike propagation are used as the final output and the gradient is calculated according to a consistent objective function $\mathcal{L}_c$:
\begin{equation}
    \mathcal{L}(x,y) = \mathcal{L}_c(\mathcal{O}_s, y) + \mathcal{L}_c(\mathcal{O}_a, y),
    \label{Eqn:final-loss}
\end{equation}
where $\mathcal{O}_s$ and $\mathcal{O}_a$ are the final outputs of auxiliary accumulation and spike propagation respectively.

\subsection{Identity Mapping}
As stated in \cite{he2016identity}, satisfying identity mapping is crucial to training a deep network. 

\paragraph{Auxiliary Accumulation Pathway.}
For auxiliary accumulation of Eq.\eqref{eqn:accumulation route}, identity mapping is achieved by when $s_l^t \equiv 0$, which can be implemented by setting the weights and the bias of the last BN layer in $f_l$ to zero.

\paragraph{The Spiking Neurons Residuals.}
The first spike propagation of Eq.\eqref{eqn:spike route} based on Spiking ResNet 
could implement identity mapping by partial spiking neurons. 
When $f_l(o_l^{t-1}) \equiv 0$, $o_l^t = {\rm SN}(o_{l}^{t-1}) \neq o_l^{t-1}$. 
To transmit $o_l^{t-1}$ and make ${\rm SN}(o_l^{t-1}) = o_l^{t-1}$, the last spiking neuron (SN) in the $l$-th residual block needs to fire a spike after receiving a spike and keep silent after receiving no spike at time-step $t$.
It works for IF neurons described by Eq.~\eqref{IFneuron}. 
Specifically, we can set $0 < V_{th} \leq 1$ and $V_{t-1}=0$ to ensure that $o_t=1$ leads to $H_t \geq V_{th}$, and $o_t=0$ leads to $H_t < V_{th}$.
Hence, we replaced all the spiking neurons at the residual connections with IF neurons.

\begin{table}[htbp]
\centering
\caption{List of element-wise functions $g$.} 
\begin{adjustbox}{max width=\linewidth}
\begin{tabular}{cl}
    \toprule
    Operator & $g(x_l^t, s_l^t)$ \\ 
    \midrule
    $\mathrm{AND}$ & $s_l^t \land x_l^t$ = $s_l^t \cdot x_l^t$ \\ 
    $\mathrm{IAND}$ & $(\neg s_l^t) \land x_l^t$ = $ (1 - s_l^t) \cdot x_l^t$ \\ 
    $\mathrm{OR}$ & $s_l^t \lor x_l^t$ = $s_l^t + x_l^t - (s_l^t \cdot x_l^t)$ \\ 
    $\mathrm{XOR}$ & $ s_l^t \oplus x_l^t$ = $ s_l^t \cdot (1 - x_l^t) + x_l^t \cdot (1 - s_l^t)$ \\ 
    \bottomrule
\end{tabular}
\end{adjustbox}
\label{tab:g}
\end{table}

\paragraph{The Element-wise Logical Residuals.}
For the second spike propagation of Eq.\eqref{eqn:spike route}, different element-wise functions $g$ in Table~\ref{tab:g} satisfy identity mapping. 
Specifically, for $\mathrm{IAND}$, $\mathrm{OR}$ and $\mathrm{XOR}$ as element-wise functions $g$, identity mapping could be achieved by setting $s_l^t \equiv 0$. 
Then $o_l^t = g(s_l^t, o_{l-1}^t) = g({\rm SN}(0), o_{l-1}^t) = g(0, o_{l-1}^t) = o_{l-1}^t$. 
This is consistent with the conditions for auxiliary accumulation Eq.\eqref{eqn:accumulation route} to achieve identity mapping and is applicable to all neuron models.
In contrast, for $\mathrm{AND}$ as the element-wise function $g$, $s_l^t$ should be ones to get identity mapping. 
Then $o_l^t = 1 \land o_{l-1}^t = o_{l-1}^t$. 
Although the input parameters of spiking neuron models can be adjusted to practice identity mapping, it is conflicted with the auxiliary accumulation pathway for identity mapping.
This is not conducive to maintaining the consistency of signal propagation between the two pathways, thus affecting the effects of training and final recognition.
Meanwhile, it is hard to control some spiking neuron models with complex neuronal dynamics to generate spikes at a specified time step.

\subsection{Vanishing Gradient Problem}


\paragraph{Auxiliary Accumulation Pathway.}
The gradient for the auxiliary accumulation pathway is calculated as $\frac{\partial a_{l+k-1}^t}{\partial a_{l}^{t}}  = \frac{\partial a_{l}^t}{\partial a_{l}^{t}}  = 1$ with identity mapping.
Since the above gradient is a constant, the auxiliary accumulation path of Eq.~\eqref{eqn:accumulation route} could also overcome the vanishing gradient problems.

\paragraph{The Spiking Neurons Residuals.}
Moreover, consider a Spiking ResNet with AAP, the gradient of the output of the $(l+k-1)$-th residual block with respect to the input of the $l$-th residual block with identity mapping could be calculated:
\begin{equation}
\label{eqn:spiking resent}
 \begin{aligned}
 \frac{\partial o_{l+k-1}^t}{\partial s_{l}^{t}} &+ \frac{\partial a_{l+k-1}^t}{\partial a_{l}^{t}} = \prod_{i=0}^{k-1} \frac{\partial o_{l+i}^{t}}{\partial s_{l+i}^{t}} + \frac{\partial a_{l}^t}{\partial a_{l}^{t}} \\
    & = \prod_{i=0}^{k-1}\Theta'(s_{l+i}^{t} - V_{th}) + \frac{\partial a_{l}^t}{\partial a_{l}^{t}} \\
    & \to 
	\begin{cases}
		1, \text{if}~0 < \Theta'(s_{l}^{t} - V_{th}) < 1 \\
		2, \text{if}~\Theta'(s_{l}^{t} - V_{th}) = 1 \\
	\end{cases},
 \end{aligned}
\end{equation}
Since the above gradient is a constant, the Spiking ResNet with AAP could alleviate the vanishing gradient problems. 

\paragraph{The Element-wise Logical Residuals.}
When the identity mapping is implemented for the spiking propagation path of Eq.\eqref{eqn:spike route}, the gradient of the output of the $(l+k)$-th dual-stream block with respect to the input of the $l$-th DSNN block could be calculated layer by layer:
\begin{equation}
 \begin{aligned}
 \frac{\partial o_{l+k-1}^t}{\partial x_{l}^{t}}  &= \prod_{i=0}^{k}\frac{\partial g(s_{l+i}^t, x_{l+i}^t)}{\partial x_{l+i}^t} \\ &= 
	\begin{cases}
		\prod_{i=0}^{k}\frac{\partial (1 \cdot x_{l+i}^t)}{\partial x_{l+i}^t}, \text{if}~~g = \mathrm{AND} \\
		\prod_{i=0}^{k}\frac{\partial ((1 - 0) \cdot x_{l+i}^t)}{\partial x_{l+i}^t}, \text{if}~~g = \mathrm{IAND} \\
		\prod_{i=0}^{k}\frac{\partial ((1 + 0 - 0) \cdot x_{l+i}^t)}{\partial x_{l+i}^t}, \text{if}~~g = \mathrm{OR} \\
		\prod_{i=0}^{k}\frac{\partial ((1+0) \cdot x_{l+i}^t)}{\partial x_{l+i}^t}, \text{if}~~g = \mathrm{XOR}
	\end{cases} = 1.
	\label{eqn:gradient_propagation}
 \end{aligned}
\end{equation}
The second equality holds as identity mapping is achieved by setting $s_{l+i}^t \equiv 1$ for $g=\mathrm{AND}$, and $s_{l+i}^t \equiv 0$ for $g=\mathrm{IAND/OR/XOR}$.
Since the gradient in Eq.~\eqref{eqn:gradient_propagation} is a constant, the spiking propagation path of Eq.\eqref{eqn:spike route} overcomes the vanishing gradient problems.

%% file: Main/Tex/05_experiments.tex
\section{Experiments}
\label{sec:exps}

\subsection{Computational Consumption}
To estimate the computational consumption of different SNNs, we calculate the number of computations required in terms of AC and MAC synaptic operations, respectively.
Moreover, \cite{Bodo2017Conversion} integrates batch normalization (BN) layers into the weights of the preceding layer with loss-less conversion. 
Therefore, we ignore BN operations when calculating the number of MAC operations, resulting in a more efficient inference consumption.
See Appendix \ref{appx:fuse} for details of the fusion process of convolution with BN.
To quantitatively estimate energy consumption, we evaluate the computational consumption based on the number of AC and MAC operations and the data for various operations in $45nm$ technology \cite{horowitz20141}, where $\mathrm{E}_{\mathrm{MAC}}$ and $\mathrm{E}_{\mathrm{AC}}$ are $4.6pJ$ and $0.9pJ$ respectively.
Here, we calculate the \textbf{D}ynamic \textbf{C}onsumption (DC) of the SNN by Eq.\eqref{eqn:energy} based on its spike firing rate on the target dataset.
Moreover, we use the \textbf{E}stimated \textbf{C}onsumption (EC) to estimate the theoretical consumption range from $[0, 100\%]$ spike firing rate, that is
\begin{align}
 \mathrm{E}(\mathcal{F}) \in [T \cdot E_{mac} \cdot O_{mac}, T \cdot (E_{ac} \cdot O_{ac}+E_{mac} \cdot O_{mac})].
\end{align}

\subsection{ImageNet Classification}
We validate the effectiveness of our Dual-stream Training method on image classification of ImageNet~\cite{deng2009imagenet} dataset. 
The IF neuron model is adopted for the static ImageNet dataset.
For a fair comparison, all of our training parameters are consistent with SEW ResNet \cite{fang2021deep}.
As shown in Table~\ref{tab:sotaimagenet}, two variations of our DST are evaluated: ``FSNN (DST)'' means the FSNN is trained by the proposed DST and AAP is removed  in the test phase, and ``DSNN'' means the APP is retained to improve the discrimination of features. 
``FSNN-18'' and "DSNN-18" represents the SNN is designed based on ResNet-18, and so on. Three types of methods are compared respectively: ``A2S'' represents ANN2SNN methods, ``FSNN'' and ``MPSNN'' means Full Spike Neural Networks and Mixed-Precision Spike Neural Networks respectively. These notations keep the same in the below.

\begin{table*}[!htb]
    \centering
    \caption{Comparison with previous Spiking ResNet on ImageNet. $\dagger$ denotes the estimated dynamic consumption based on the spike firing rate provided in the corresponding paper. 
    A2S represents ANN2SNN methods, FSNN and MPSNN mean Full Spike Neural Networks and Mixed-Precision Spike Neural Networks respectively. FSNN (DST) represent the FSNN is trained by the proposed DST and DSNN means APP is retained in the test phase.
    }
    \begin{adjustbox}{max width=\linewidth}
        \begin{tabular}{lccccccc}
            \toprule
            Network & Methods & Acc@1 & $T$ & EC(mJ) & $O_{ac}$(G) & $O_{mac}$(G) & DC(mJ)\\
            \midrule
            PreAct-ResNet-18 \cite{meng2022training} & A2S & 67.74 & 50 &  $[1.43,77.84]$ & - & -  & -\\
            Spiking ResNet-34 \cite{rathi2020enabling} & A2S & 61.48 & 250 &  $[7.31,805.79]$ & - & -  & - \\
            Spiking ResNet-34 \cite{pmlr-v139-li21d} & A2S & 74.61 & 256 & $[7.45,825.29]$ & - & -  & - \\
            \hline
            Spiking ResNet-34 with td-BN \cite{zheng2020going} & FSNN & 63.72  & 6 & $[0.69,19.85]$ & 5.34 & 0.15 & 5.50 $^{\dagger}$\\
            Spiking ResNet-50 with td-BN \cite{zheng2020going} & FSNN & 64.88  & 6 & $[1.10,22.59]$ & 6.01 & 0.24 & 6.52 $^{\dagger}$\\
            Spiking ResNet-34~\cite{deng2021temporal} & FSNN & 64.79 & 6 & $[0.69,19.85]$ & 5.34 & 0.15 & 5.50 $^{\dagger}$ \\ 

            FSNN-18 (DST) & FSNN & 62.16 & 4 & $[0.55,4.31]$ & 1.69 & 0.12 & 2.07 \\
            FSNN-34 (DST) & FSNN & 66.45 & 4 & $[0.55,7.64]$ & 3.42 & 0.12 & 3.63 \\
            FSNN-50 (DST) & FSNN & 67.69 & 4 & $[0.55,10.42]$ & 3.14 & 0.12 & 3.38 \\
            FSNN-101 (DST) & FSNN & 68.38 & 4 & $[0.55,20.64]$ & 4.42 & 0.12 & 4.53 \\
            
            \midrule
            SEW ResNet-18 (ADD) \cite{fang2021deep} & MPSNN & 63.18 & 4 &  $[12.65,16.40]$ & 0.51 & 2.75 & 13.11\\
            SEW ResNet-34 (ADD) \cite{fang2021deep} & MPSNN & 67.04 & 4 &  $[29.72,36.78]$ & 0.86 & 6.46 & 30.50\\
            SEW ResNet-50 (ADD) \cite{fang2021deep} & MPSNN & 67.78 & 4 &  $[23.64,33.50]$ & 2.01 & 5.14 & 25.45 \\
            SEW-ResNet-34 (ADD) \cite{deng2021temporal} & MPSNN & 68.00 & 4 & $[29.72,36.78]$ & 0.86 & 6.46 & 30.50 $^{\dagger}$ \\
            SEW ResNet-101 (ADD) \cite{fang2021deep} & MPSNN & 68.76 & 4 & $[39.88,59.97]$ & 3.07 & 8.67 & 42.65 \\
            DSNN-18 & MPSNN & 63.46 & 4 & $[0.92,4.67]$ & 1.69 & 0.20 & 2.44 \\
            DSNN-34 & MPSNN & 67.52 & 4 & $[0.92,8.00]$ & 3.42 & 0.20 & 4.00 \\
            DSNN-50 & MPSNN & 69.56 & 4 & $[6.30,16.17]$ & 3.20 & 1.37 & 9.18 \\
            DSNN-101 & MPSNN & 71.12 & 4 & $[6.30,26.39]$ & 4.48 & 1.37 & 10.33 \\
            \hline
            \hline
        \end{tabular}
    \end{adjustbox}
    \label{tab:sotaimagenet}
    \vspace{-0.4cm}
\end{table*}

As shown in Table \ref{tab:sotaimagenet}, we can obtain 3 following key findings: 

First, the SOTA ANN2SNN methods \cite{pmlr-v139-li21d, hu2020spiking} achieve higher accuracies than FSNN as well as other SNNs trained from scratch, but they use 64 and 87.5 times as many time-steps as FSNN respectively, which means that they require more computational consumption.
Since most of these methods do not provide the trained models and their DCs are not available, we only used EC to evaluate the consumption. 
From Table~\ref{tab:sotaimagenet}, these models with larger time steps also have larger computational consumption.
\cite{meng2022training} also achieves good performance with ResNet-18, but its SNN acquisition process is relatively complex. It  needs ANN to pre-train the model first and  then retrain the SNN. 
Although its time steps are less than other ANN2SNN methods, it is still 12.5 times that of DSNN. Due to ANN2SNN being a different type of method from ours, the comparisons are listed just for reference.  

Second, for full-spike neural networks, FSNN (DST) outperforms Spiking ResNet~\cite{zheng2020going,deng2021temporal} even with lower computational consumption.
The performance of FSNN (DST) has a major advantage over other FSNNs and also improves with the increasing depth of the network. It indicates that the proposed DST is indeed helpful to train FSNN.

Finally, the performance of DSNN is further improved when AAP is added to FSNN at the inference phase.
At the same time, the DSNN offers superior performance compared to the mixed-precision SEW ResNet~\cite{fang2021deep}.
Note that the performance gap between SEW ResNet-34 and SEW ResNet-50 is not big, and the computational consumption of SEW ResNet-34 is higher than SEW ResNet-50. 
This phenomenon comes from that ResNet-34 and ResNet-50 use BasicBlock \cite{he2015deep} and Bottleneck \cite{he2015deep} as blocks respectively. 
As shown in Figure~\ref{fig:blocks} (b), the first-layer convolution of each block is regarded as a MAC computation operation due to the non-spike data generated by ADD function. 
However, the first-layer convolution of BasicBlock is much larger than the synaptic operation of Bottleneck, which causes the computational consumption of SEW ResNet-34 to be greater than that of ResNet-50. In contrast, DSNN ensures that the input and output of each block are spike data through the dual-stream mechanism, so that the theoretical computational consumption increases linearly with the increase of network depth. 


\subsection{DVS Classification}

\paragraph{DVS Gesture.}
We also compare our method with SEW ResNet-7B-Net \cite{fang2021deep} on the DVS Gesture dataset \cite{amir2017low}, which contains 11 hand gestures from 29 subjects under 3 illumination conditions. 
We use the similar network structure 7B-Net in \cite{fang2021deep}. 
As shown in Table~\ref{tab:dvs}, FSNN-7 (DST) with $\mathrm{IAND}$ obtains better performance than other FSNN methods, such as SEW with $\mathrm{IAND}$ and td-BN, demonstrating the DST is effective to train FSNN.
In addition, even though SEW utilizes $\mathrm{IAND}$ as  $g(\cdot)$ which brings non-spike operations, our  FSNN-7 (DST)  still achieves comparable performance, and only requires a tenth of the computational consumption of SEW.

\paragraph{CIFAR10-DVS.}
We also evaluate Spiking ResNet models on the CIFAR10-DVS dataset \cite{cifar10_dvs}, which is obtained by recording the moving images of the CIFAR-10 dataset on an LCD monitor by a DVS camera. 
We use Wide-7B-DSNN which is a similar network structure Wide-7B-Net in \cite{fang2021deep}. 
As shown in Table~\ref{tab:dvs}, FSNN-7 (DST)  achieves better performance and lower computational consumption than the previous Spiking ResNet \cite{zheng2020going} and SEW ResNet \cite{fang2021deep}.

\begin{table}[!tb]
    \caption{Comparison with Spiking ResNet methods on DVS Gesture and CIFAR10-DVS. Once the ADD function is used as $g(\cdot)$, it will bring non-spiking operation to SEW~\cite{fang2021deep}. The comparison with it is listed just for reference.}
    \centering
    \label{tab:dvs}
    \begin{adjustbox}{max width=\linewidth}
    \begin{tabular}{lcccc}
        \toprule
        Networks & $g(\cdot)$ & DVS Gesture & CIFAR10-DVS & $T$ \\ 
        & & \multicolumn{2}{c}{ACC/DC(mJ)} \\
        \midrule
         SEW~\cite{fang2021deep} & $\mathrm{ADD}$ & \textbf{97.92}/17.09 & 74.4/16.71 & 16 \\
        \midrule
        SEW~\cite{fang2021deep} & $\mathrm{IAND}$ & 95.49/1.48 & - & 16 \\
        td-BN~\cite{zheng2020going} & - & 96.87/- & 67.8/- & 40/10 \\
        \midrule
        FSNN-7 (DST) & $\mathrm{AND}$ & 55.56/2.59 & 70.9/3.57 &  16 \\
        FSNN-7 (DST)& $\mathrm{OR}$ & 96.18/1.04 & 74.8/3.51 & 16 \\
        FSNN-7 (DST)& $\mathrm{XOR}$ & 96.53/1.19 & 74.0/3.41 & 16 \\
        FSNN-7 (DST)& $\mathrm{IAND}$ & 97.57/1.10 & 73.1/4.65 & 16 \\
        \bottomrule
    \end{tabular}
    \end{adjustbox}
\end{table}

\subsection{Further Analysis}

\paragraph{Computational Consumption}
Here, we analyze the computational consumption advantages of DSNN.
First, the networks in most ANN2SNN  methods are full-spike, where few MAC operations are from batch normalization and conversion of images to spikes.
However, their computational consumption is still very large due to the large time steps.
DSNN and FSNN has fewer time steps than ANN2SNN methods, which means that its theoretical maximum consumption is much smaller than ANN2SNN as shown in Table~\ref{tab:sotaimagenet}.
%
Second, the DSNN has lower EC and DC than Spiking ResNet based on addition (SEW ResNet), and achieves better performance.
The additive-based SEW ResNet will increase its AC and MAC as the network gets deeper, which will bring about a multifold increase in computational consumption.
In contrast, the main MAC operation for DSNN comes from the downsampling process of accumulated signals in spike accumulation.
Therefore, a deeper DSNN only increases the AC operation, and its MAC operation is constant as shown in Table~\ref{tab:sotaimagenet}.
Finally, it also reveals for the first time the positive relationship between computational consumption and the performance of SNNs.
In addition, the performance of DSNN increases gradually with the increase in computational consumption. It indicates the added computational consumption of deeper DSNN is meaningful.

\begin{figure}[tbp]
    \centering
    \setlength{\abovecaptionskip}{0.cm}
    \includegraphics[width=\linewidth]{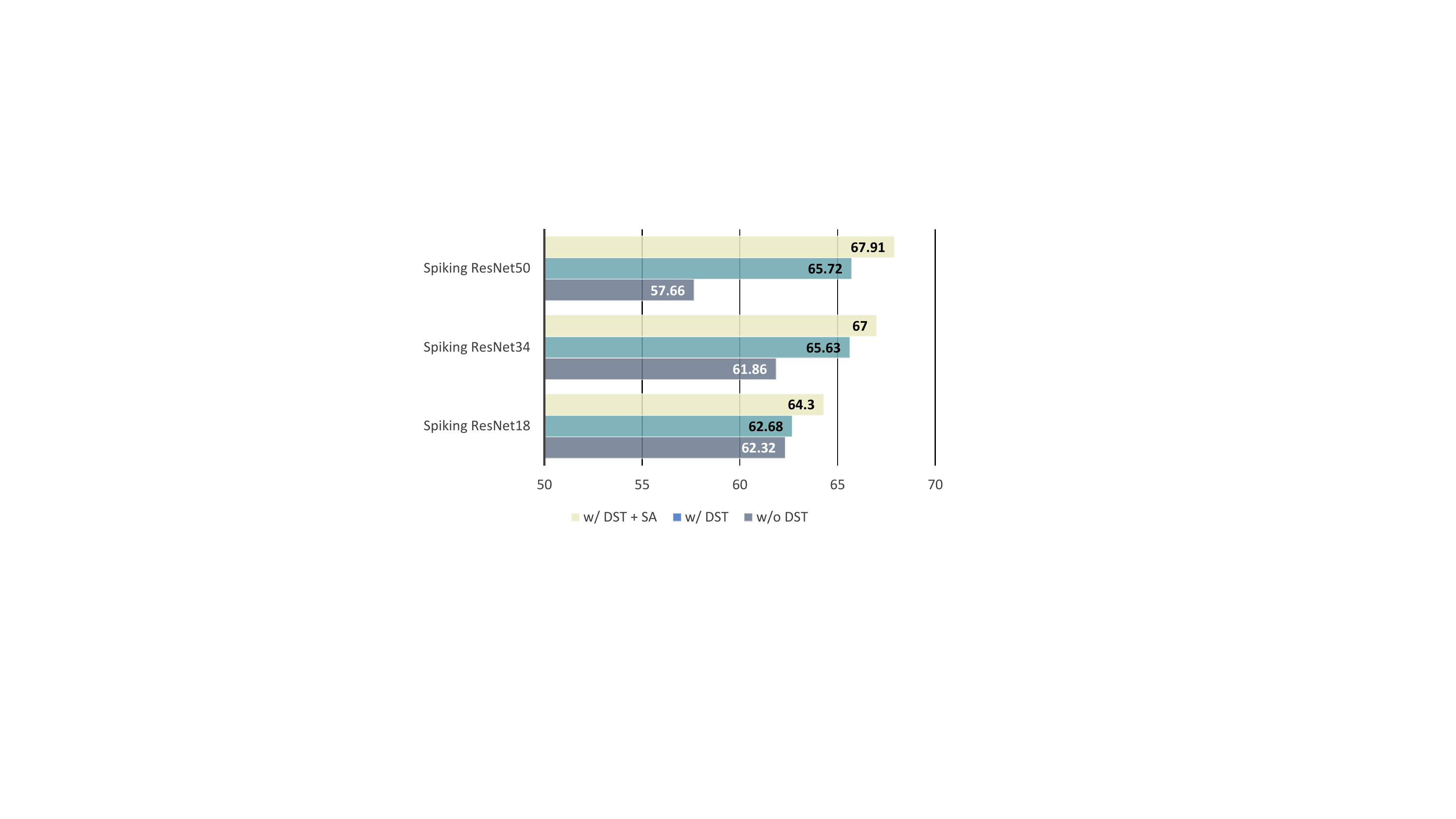}
    \vspace{-0.5cm}
    \caption{Training Spiking ResNet from scratch on ImageNet with/without Dual-Stream Training.}
    \label{fig:spike resnet}
    \vspace{-0.5cm}
\end{figure}

\paragraph{Vanishing Gradient Problem.}

As mentioned in Section~\ref{sec:spiking resnet}, Spiking ResNet suffers from a vanishing gradient problem. As shown in Figure~\ref{fig:spike resnet}, the performance of the Spiking ResNet gradually decreases as the number of network layers increases. Network depth did not bring additional gain to the original Spiking ResNet. 
With the addition of DST, the performance of the Spiking ResNet gradually increases with increasing network depth. 
Also, the performance is superior with the addition of AAP. 
As demonstrated in Eq.\eqref{eqn:spiking resent}, DST could help alleviate the vanishing gradient problem of the Spiking ResNet.

\begin{figure}[htbp]
	\centering
	\subfigure[FSNN (DST)]{
		\begin{minipage}[t]{0.43\linewidth}
			\centering
			\includegraphics[width=0.8\linewidth]{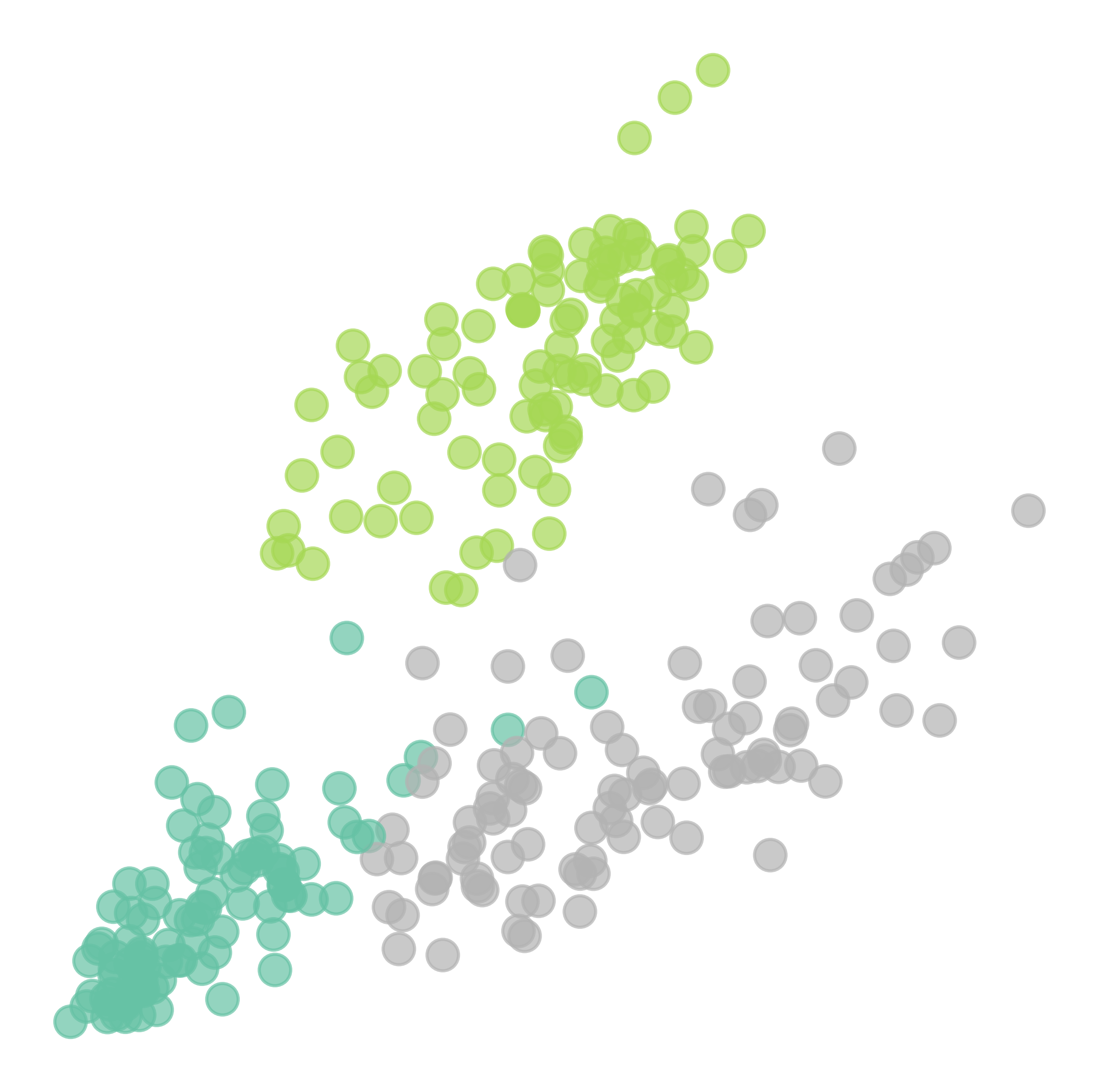}
		\end{minipage}%
	}
	\subfigure[DSNN]{
		\begin{minipage}[t]{0.43\linewidth}
			\centering
			\includegraphics[width=0.8\linewidth]{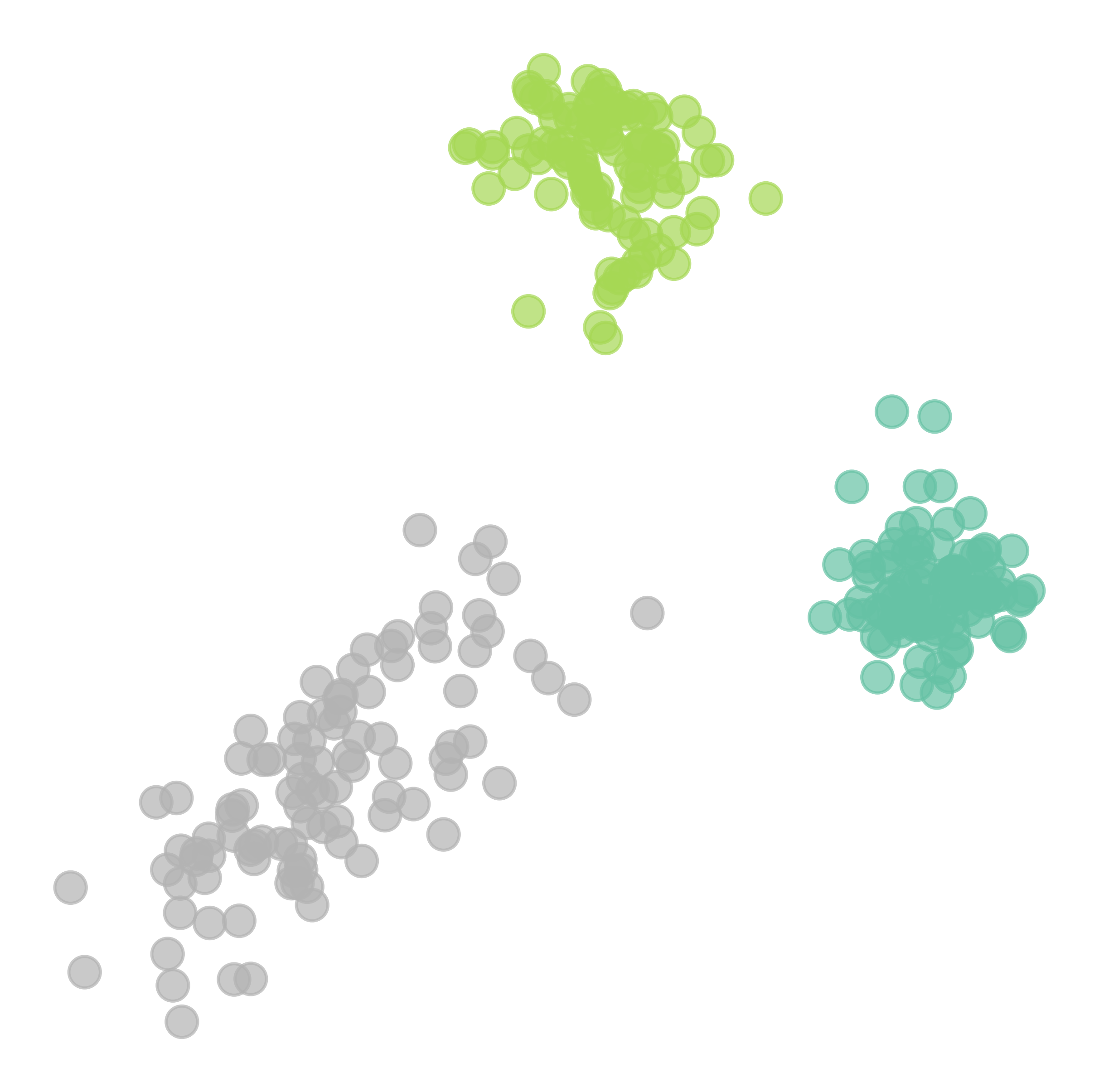}
		\end{minipage}%
	}
	\caption{The the t-SNE \cite{maaten2008visualizing} plots of embedding features on DVS Gesture.}
	\label{fig:tsne}
    \vspace{-0.4cm}
\end{figure}

\paragraph{Feature Enhancement.}
Here, we analyze the effects of auxiliary accumulation in terms of improving feature discrimination. 
Moreover, AAP increases the feature discriminant of forward as shown in Figure~\ref{fig:tsne}.
Under fewer time steps, the spike feature is almost a binary feature, because its discrete property limits the discriminant of its feature.
The discrimination of features from spike propagation is not enough, which affects the overall performance.
On the other hand, we could increase the number of deep spike features by increasing time steps, so as to improve the discrimination by increasing the number of features, which often leads to too high computational consumption as shown in Table~\ref{tab:sotaimagenet}.
More importantly, AAP separates the computation of AC operations from the MAC computation of recognition, allowing SNN to take full advantage of its low computational consumption.

\paragraph{Spike Transformer.}
We also test the role of DST on the Transformer on CIFAR100. The latest method Spikeformer \cite{zhou2023spikformer} utilizes $\mathrm{ADD}$ as its residual connection and achieve SOTA performance. However, the $\mathrm{ADD}$ introduces additional computational consumption from non-spike data. As shown in Table~\ref{tab:spikeformer}, once the $\mathrm{IAND}$ is replaced by $\mathrm{IAND}$, the  performance of the full-spike Spikeformer will drop. After introducing APP as shown in Figure~\ref{fig:blocks}, the Spikeformer with a full-spike signal achieves similar performance to the original mixed-precision Spikeformer. Notably, there is no downsampling in the Spikeformer, which further exploits the advantages of AAP while not introducing additional MAC operations. 
The small amount of MAC operations comes mainly from the image-to-spike conversion and the classifier operations.


\begin{table}[htbp]
    \caption{Learning Spike Transformer~\cite{zhou2023spikformer} with DST. ADD will bring additional computational consumption from non-spike data. Spikformer (IAND) is the full-spike version of Spikformer, and Spikformer (IAND) w/ DST means the full-spike Spikformer is trained by our DST.}
    \centering
    \label{tab:spikeformer}
    \begin{adjustbox}{max width=\linewidth}
    \begin{tabular}{lcccc}
        \toprule
        Networks & Method & Acc & DC(mJ) \\ 
        \midrule
        Spikformer (ADD) & MPSNN & 77.21 & 5.27 \\
        \midrule
        Spikformer (IAND) & FSNN & 75.54 & 0.82 \\
        Spikformer (IAND) w/ DST & FSNN & 76.96 & 0.84 \\
        \bottomrule
    \end{tabular}
    \end{adjustbox}
\end{table}

%% file: Main/Tex/06_conclusion.tex
\section{Conclusion \& Outlook}

In this paper, we point out that the main contradiction of FSNNs comes from information loss of full spike propagation.
Therefore, we propose the Auxiliary Accumulation Pathway with consistent identity mapping which is able to compensate for the loss of information forward and backward from full spike propagation, so as to keep computationally efficient and high-performance recognition simultaneously.
The experiments on the ImageNet, DVS Gesture, and CIFAR10-DVS indicate that DST could improve the ResNet-based and Transformer-baed SNNs trained from scratch in both accuracy and computation consumption.
Looking forward, this SNN with a dual-stream structure may provide a reference for the design of neuromorphic hardware, which could improve computational efficiency by separating spike and non-spike computation.
In addition, the proposed dual-streams mechanism  is similar to the dual-streams object recognition pathway in the human brain, which may provide a new potential direction for SNN structure design.


%% file: Main/Tex/07_appendix.tex
\appendix


\section{The Fusion of $\mathrm{Conv}+\mathrm{BN}$}
\label{appx:fuse}
\textbf{The fusion of convolution and batch normalization}.
It should be noted that the spike signal becomes the floating point once it has passed through the convolution layer, which leads to subsequent MAC operations from the Batch Normalization (BN).
However, the homogeneity of convolution allows the following BN and linear scaling transformation to be equivalently fused into the convolutional layer with an added bias~\cite{ding2019acnet,ding2021repvgg}.
Specifically, each BN and its preceding convolution layer into a convolution with a bias vector. 
Let $\{ \mathrm{W}^\prime,\mathrm{B}^\prime \}$ be the kernel and bias converted from $\{\mathrm{W},\vect{\mu},\vect{\sigma},\vect{\gamma},\vect{\beta}\}$, we have
\begin{equation}
\label{eq-fuse-bn}
\mathrm{W}^\prime = \frac{\vect{\gamma}_i}{\vect{\sigma}_i}\mathrm{W} \,,\quad \mathrm{B}^\prime_i = -\frac{\vect{\mu}_i \vect{\gamma}_i}{\vect{\sigma}_i} + \vect{\beta}_i \,.
\end{equation}
Then it is easy to verify that,
\begin{equation}
    \text{bn}(\mathrm{M}\ast\mathrm{W},\vect{\mu},\vect{\sigma},\vect{\gamma},\vect{\beta}) = (\mathrm{X} \ast \mathrm{W}^\prime) + \mathrm{B}^\prime \,.
    \label{Eqn:fusion}
\end{equation}
Therefore, when calculating the theoretical computational consumption, ignore the consumption of the BN could be ignored.

\section{Implementation Details}
\label{sec:id}
All experiments are implemented with SpikingJelly~\cite{SpikingJelly}, which is an open-source deep learning framework for SNNs based on PyTorch~\cite{PYTORCH}. The source code is included in the supplementary. 
The hyper-parameters of the DSNN for different datasets are shown in Table~\ref{tab:hyp}. 
The pre-processing data methods for three datasets are as follows:

\begin{table}[htbp]
	\centering
        \caption{Hyper-parameters of the DSNN for ImageNet, DVS Gesture, and CIFAR10-DVS datasets. CA denotes the Cosine Annealing \cite{loshchilov2016sgdr}.}
        \begin{adjustbox}{max width=\linewidth}
		\begin{tabular}{llllll}
                \toprule
			Dataset & Learning Rate Scheduler & Epochs & $ \boldsymbol {lr}$ & Batch Size & $\boldsymbol {T}$ \\
			\midrule
			ImageNet & CA, $T_{max}=320$ & 320 &0.1 & 128 & 4 \\
			DVS Gesture & Step, $T_{step}=64. \gamma=0.1$ & 192 & 0.005 & 16 & 16\\
			CIFAR10-DVS & CA, $T_{max}=64$ & 64 & 0.01 & 16 & 16\\
                CIFAR100 & CA & 310 & $5e-4$ & 128 & 4 \\ 
			\bottomrule
		\end{tabular}
        \end{adjustbox}
	\label{tab:hyp}
\end{table}

\noindent
\textbf{ImageNet.} 
The data augmentation methods used in \cite{he2015deep} are also applied in our experiments. A 224×224 crop is randomly sampled from an image or its horizontal flip with data normalization for train samples. A 224×224 resize and central crop with data normalization are applied for test samples. 

\noindent
\textbf{DVS Gesture.} 
We utilize \textit{random temporal delete}~\cite{fang2021deep} to relieve overfitting. Denote the sequence length as $T$, we randomly delete $T - T_{train}$ slices in the original sequence and use $T_{train}$ slices during training. During inference we use the whole sequence, that is, $T_{test} = T$. We set $T_{train}=12, T=16$ in all experiments on DVS Gesture.

\noindent
\textbf{CIFAR10-DVS.} 
We use the AER data pre-processing in \cite{fang2020incorporating} for DVS128 Gesture. We do not use \textit{random temporal delete} because CIFAR10-DVS is obtained by static images.

\noindent
\textbf{CIFAR100.} 
All hyperparameters are consistent with \cite{zhou2023spikformer}.

\begin{figure}[htbp]
    \centering
    \setlength{\abovecaptionskip}{0.cm}
    \includegraphics[width=0.7\linewidth]{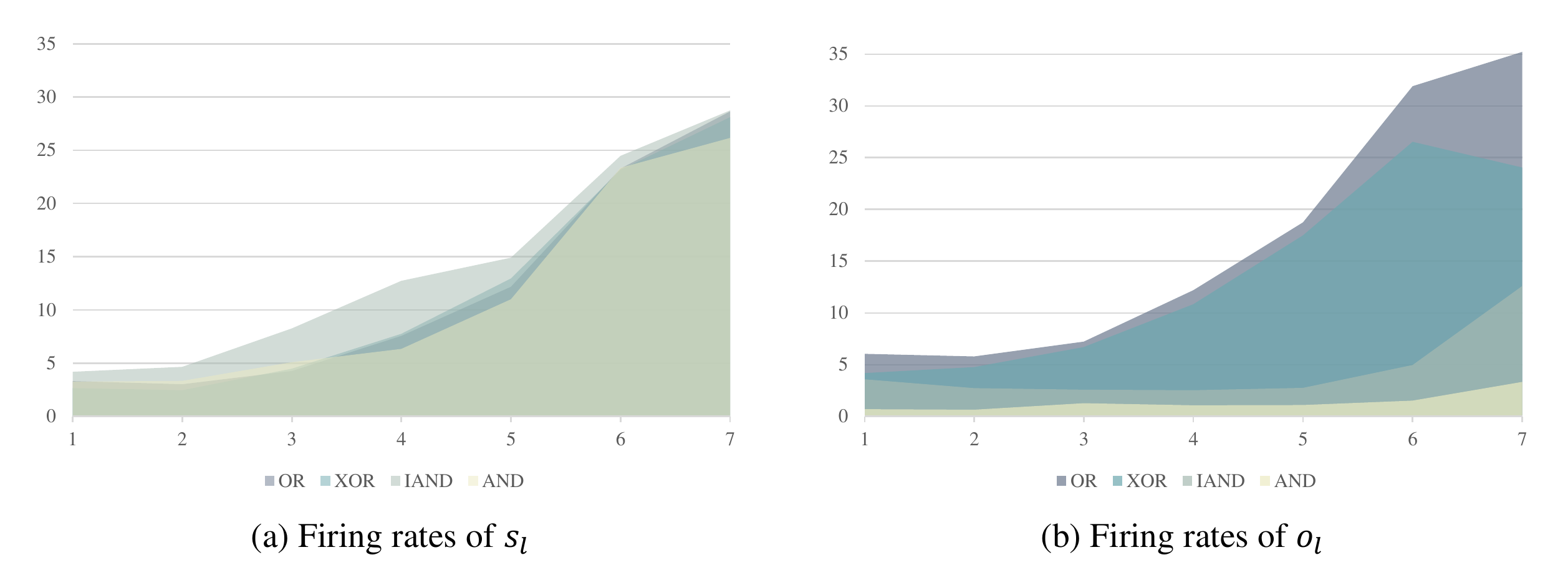}
    \caption{Firing rates of Dual-stream Blocks on DVS Gesture.}
    \label{fig:fires}
\end{figure}

\section{More Analysis}

\subsection{Analysis of Spiking Response of Dual-Stream Block}
Furthermore, we analyze the firing rates of DSNN, which are closely related to computational power consumption.
Figure~\ref{fig:fires}(a) shows the firing rates of $o_l$ for the Dual-Stream Block on DVS Gesture, where all spiking neurons in dual-stream blocks have low firing rates, and the spiking neurons in the first two blocks even have firing rates of almost zero.
All firing rates in the dual-stream blocks are not larger than 0.5, indicating that all neurons fire on average not more than two spikes. 
The firing rates of $o_l$ in the first few blocks are at a low level, verifying that most dual-stream blocks act as identity mapping.
As the depth of the network accelerates, the fire rate increases, increasing the ability to express features for better recognition.

\subsection{Evaluation of Different Element-wise Functions}
We evaluate all kinds of element-wise functions $g$ on CIFAR10-DVS and DVS Gesture as shown in Table~\ref{tab:dvs}. 
As mentioned above, both pathways of DSNN should achieve identity mapping under the same conditions, and  $\mathrm{IAND}$, $\mathrm{OR}$, and $\mathrm{XOR}$  (except $\mathrm{AND}$) satisfy  this. 
As shown in Table~\ref{tab:dvs}, the performance of $\mathrm{AND}$ is lower than other functions as expected. It also indicates the necessity of identity mapping.
Moreover, $\mathrm{IAND}$, $\mathrm{OR}$, and $\mathrm{XOR}$ all achieved relatively good performance on CIFAR10-DVS and DVS Gesture. 
On the other hand, Figure~\ref{fig:fires} shows the fire rates for different element-wise $g(\cdot)$ functions. 
For different element-wise $g(\cdot)$ functions, the output $s_l$ of each block is not obviously different. 
However, for the fire rate after the element-wise $g(\cdot)$ function, the output $o_l$ gap of each block is obvious. 
As shown in Figure~\ref{fig:fires}(b), $\mathrm{OR}>\mathrm{XOR}>\mathrm{IAND}>\mathrm{AND}$ for the firing rate. 
For datasets with different complexity, different firing rates can be needed for better recognition. 
For example, $\mathrm{IAND}$ achieves the best performance for a simple DVS dataset. 
However, for the slightly complex CIFAR10-DVS, $\mathrm{OR}$  has a higher rate to improve the feature expression ability, thus achieving the best performance.

\subsection{Gradients Check on Deeper DSNN}

Eq.~(11) and Eq.~(12) analyze the gradients of multiple blocks with identity mapping. To verify that DSNN can overcome the vanishing/exploding gradient, we check the gradients of the deeper DSNN with 50 layers. 
In this paper, the surrogate gradient method~\cite{neftci2019surrogate} is used to define $\Theta'(x) \triangleq \sigma'(x)$ during error back-propagation, with $\sigma(x)$ denote the surrogate function.
The surrogate gradient function we used in all experiments is $\sigma(x) = \frac{1}{\pi} \arctan(\frac{\pi}{2}\alpha x) + \frac{1}{2}$, thus $\sigma'(x) = \frac{\alpha}{2(1 + (\frac{\pi}{2} \alpha x)^2)}$.

\begin{figure}[htbp]
	\centering
	\subfigure[Firing rate of $o_l$]{
		\begin{minipage}[t]{0.35\linewidth}
			\centering
			\includegraphics[width=\linewidth]{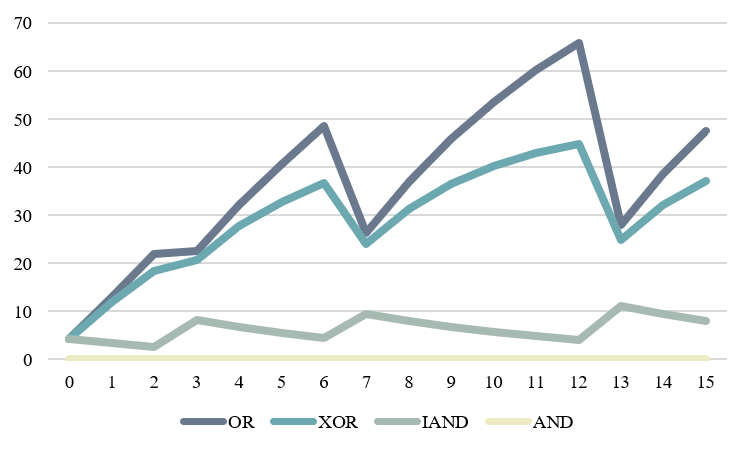}
			\label{fig:fio}
		\end{minipage}%
	}
	\subfigure[Firing rate of $s_l$]{
		\begin{minipage}[t]{0.35\linewidth}
			\centering
			\includegraphics[width=\linewidth]{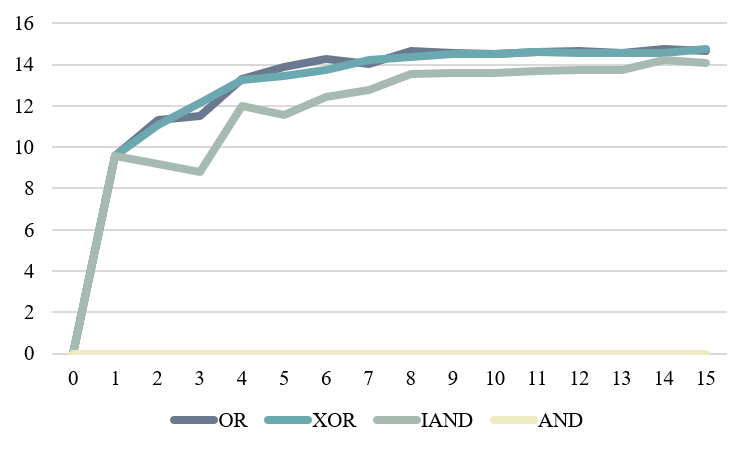}
			\label{fig:fis}
		\end{minipage}%
	}
	\caption{The initial firing rates of output $o_l$ and $s_l$ in $l$-th block on 50 layer network.}
	\label{fig:firesinit}
\end{figure}

\paragraph{Initial Firing Rates}
As the gradients of SNNs are significantly influenced by initial firing rates~\cite{fang2021deep}, he we analyze the firing rate.
Figure~\ref{fig:firesinit} shows the initial firing rate of $l$-th block's output $o_l$, where the mutations occur due to downsampling blocks.
As shown in Figure~\ref{fig:fio}, the silence problem happens in the DSNN with $\mathrm{AND}$ (yellow curve).
When using $\mathrm{AND}$, $o_l^t = {\rm SN}(f^{l}(o_{l-1}^t)) \land o_{l-1}^t \leq o_{l-1}^t$. Since it is hard to keep ${\rm SN}(f^{l}(o_{l-1}^t)) = 1$ at each time-step $t$, the silence problem may frequently happen in DSNN with $\mathrm{AND}$.
In contrast, compared with $\mathrm{AND}$, using $\mathrm{OR}$, $\mathrm{XOR}$ and $\mathrm{IAND}$ could easily maintain a certain firing rate.
Figure~\ref{fig:fis} shows the firing rate of $s_l = {\rm SN}(f^{l}(o_{l-1}^t))$, which represents the output of the last SN in $l$-th block. 
It shows that although the firing rate of $o_l$ in DSNN with $\mathrm{OR}$, $\mathrm{XOR}$ and $\mathrm{IAND}$ could increase constantly with the depth of networks in theory, the last SN in each block still maintains a stable firing rate in practice. 

\begin{figure*}[htbp]
	\centering
	\subfigure[$\mathrm{OR}$]{
		\begin{minipage}[b]{0.23\linewidth}
			\centering
			\includegraphics[width=\linewidth]{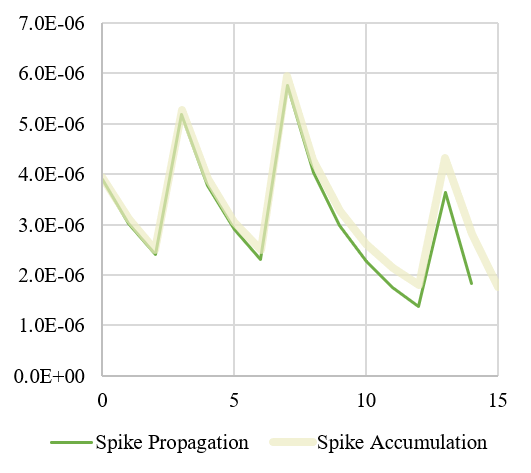}
			\label{fig:org2}
		\end{minipage}%
	}
	\subfigure[$\mathrm{XOR}$]{
		\begin{minipage}[b]{0.23\linewidth}
			\centering
			\includegraphics[width=\linewidth]{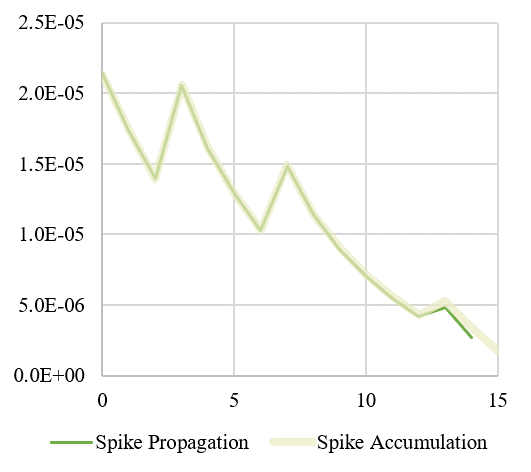}
			\label{fig:xorg2}
		\end{minipage}%
	}
	\subfigure[$\mathrm{IAND}$]{
		\begin{minipage}[b]{0.23\linewidth}
			\centering
			\includegraphics[width=\linewidth]{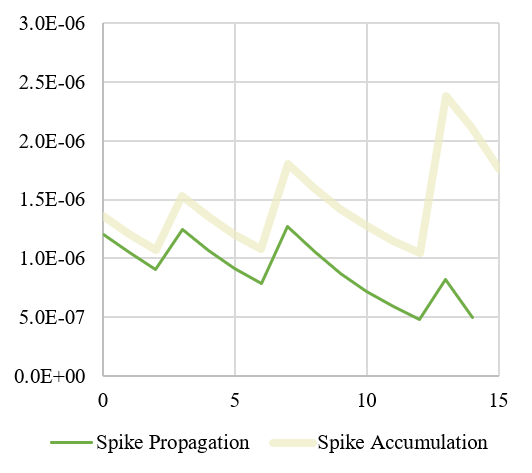}
			\label{fig:iandg2}
		\end{minipage}%
	}
	\subfigure[$\mathrm{AND}$]{
		\begin{minipage}[b]{0.23\linewidth}
			\centering
			\includegraphics[width=\linewidth]{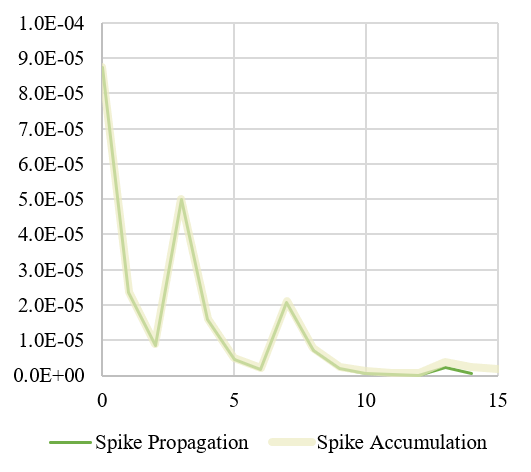}
			\label{fig:andg2}
		\end{minipage}%
	}\\
        \subfigure[$\mathrm{OR}$ w/o SA]{
		\begin{minipage}[b]{0.23\linewidth}
			\centering
			\includegraphics[width=\linewidth]{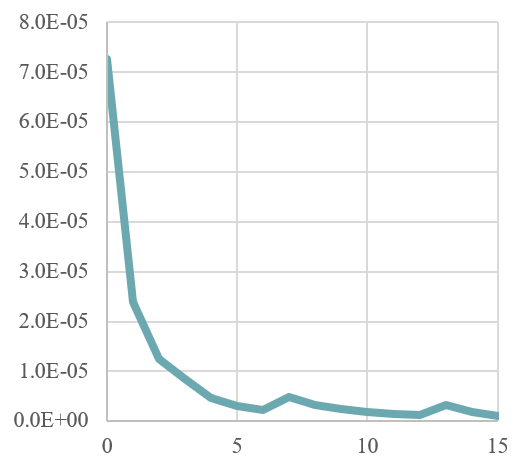}
			\label{fig:org2sa}
		\end{minipage}%
	}
	\subfigure[$\mathrm{XOR}$ w/o SA]{
		\begin{minipage}[b]{0.23\linewidth}
			\centering
			\includegraphics[width=\linewidth]{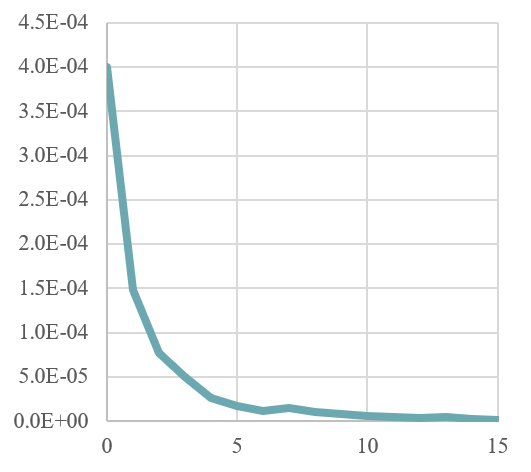}
			\label{fig:xorg2sa}
		\end{minipage}%
	}
	\subfigure[$\mathrm{IAND}$ w/o SA]{
		\begin{minipage}[b]{0.23\linewidth}
			\centering
			\includegraphics[width=\linewidth]{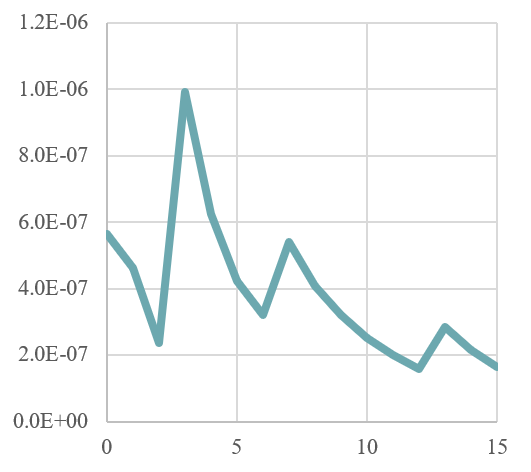}
			\label{fig:iandg2sa}
		\end{minipage}%
	}
	\subfigure[$\mathrm{AND}$ w/o SA]{
		\begin{minipage}[b]{0.23\linewidth}
			\centering
			\includegraphics[width=\linewidth]{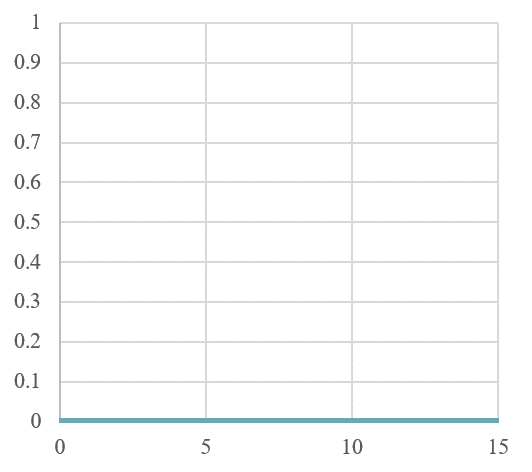}
			\label{fig:andg2sa}
		\end{minipage}%
        }
	\caption{Gradient amplitude $\left\| \frac{\partial L}{\partial s_l} \right\|$ of $l$-th block when $V_{th}=1, \alpha=2$ in the surrogate function $\sigma(x)$.}
	\label{fig:grad2}
\end{figure*}

\paragraph{Vanishing Gradient}
To analysis the vanishing gradient, we set $V_{th}=1$ and $\alpha=2$ in  the surrogate function $\sigma(x)$. In this case, $\sigma'(x) \leq \sigma'(0) = \sigma'(1 - V_{th}) = 1$ and $\sigma'(0 - V_{th}) = 0.092 < 1$, and transmitting spikes to SNs is prone to causing vanishing gradient.
The gradient amplitude $\left\| \frac{\partial L}{\partial S^{l}} \right\|$ of each block is shown in Figure~\ref{fig:grad2}(a-d), and DSNN do not be affected no matter what $g$ we choose.This is caused that in the identity mapping areas, $s_l$ is constant for all index $l$, and the gradient also becomes a constant as it will not flow through SNs.  Compared with the case where only spike propagation exists (Figure~\ref{fig:grad2}(e-h)), the gradient in DSNN is more stable. In addition, DSNN even could significantly improve $\mathrm{AND}$ which would limit the neuron firing and affect the gradient. 
It indicates the effectiveness of the DSNN for the vanishing gradient.

\begin{figure*}[htbp]
	\centering
	\setlength{\abovecaptionskip}{0.cm}
	\subfigure[$\mathrm{OR}$]{
		\begin{minipage}[b]{0.23\linewidth}
			\centering
			\includegraphics[width=\linewidth]{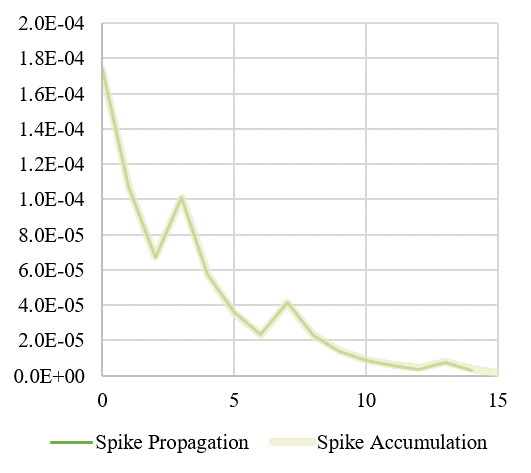}
			\label{fig:org3}
		\end{minipage}%
	}
	\subfigure[$\mathrm{XOR}$]{
		\begin{minipage}[b]{0.23\linewidth}
			\centering
			\includegraphics[width=\linewidth]{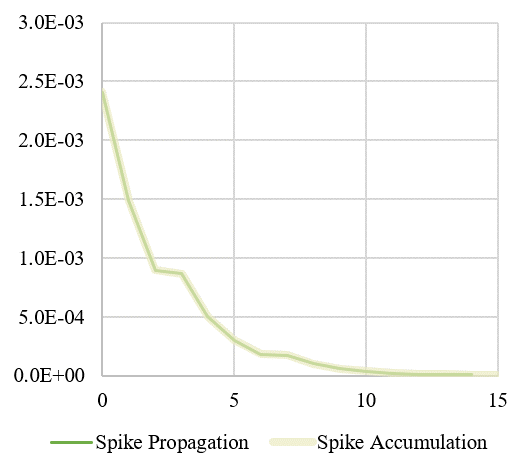}
			\label{fig:xorg3}
		\end{minipage}%
	}
	\subfigure[$\mathrm{IAND}$]{
		\begin{minipage}[b]{0.23\linewidth}
			\centering
			\includegraphics[width=\linewidth]{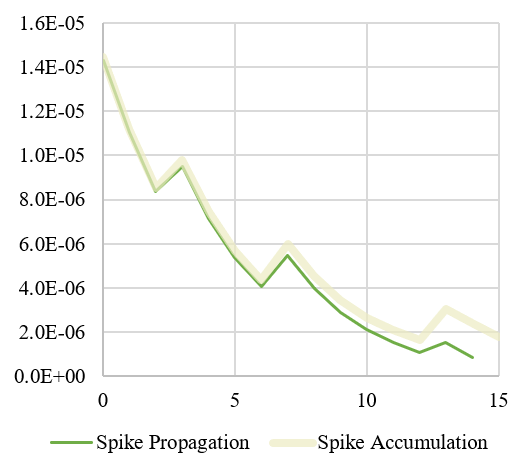}
			\label{fig:iandg3}
		\end{minipage}%
	}
	\subfigure[$\mathrm{AND}$]{
		\begin{minipage}[b]{0.23\linewidth}
			\centering
			\includegraphics[width=\linewidth]{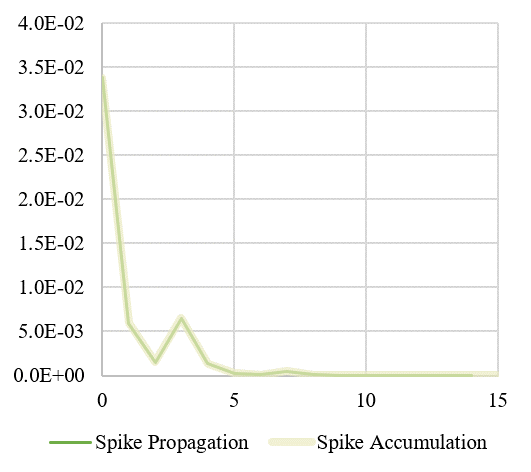}
			\label{fig:andg3}
		\end{minipage}%
	}\\
        \subfigure[$\mathrm{OR}$ w/o SA]{
		\begin{minipage}[b]{0.23\linewidth}
			\centering
			\includegraphics[width=\linewidth]{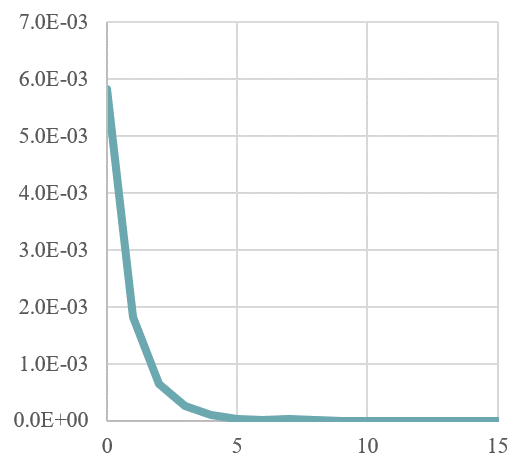}
			\label{fig:org3sa}
		\end{minipage}%
	}
	\subfigure[$\mathrm{XOR}$ w/o SA]{
		\begin{minipage}[b]{0.23\linewidth}
			\centering
			\includegraphics[width=\linewidth]{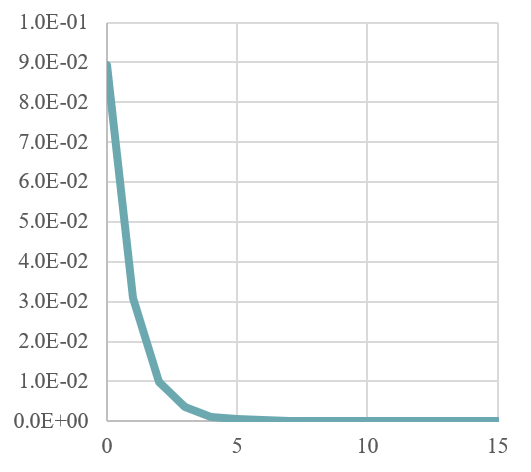}
			\label{fig:xorg3sa}
		\end{minipage}%
	}
	\subfigure[$\mathrm{IAND}$ w/o SA]{
		\begin{minipage}[b]{0.23\linewidth}
			\centering
			\includegraphics[width=\linewidth]{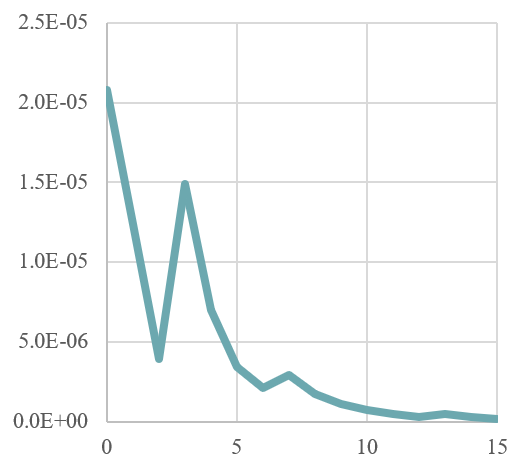}
			\label{fig:iandg3sa}
		\end{minipage}%
	}
	\subfigure[$\mathrm{AND}$ w/o SA]{
		\begin{minipage}[b]{0.23\linewidth}
			\centering
			\includegraphics[width=\linewidth]{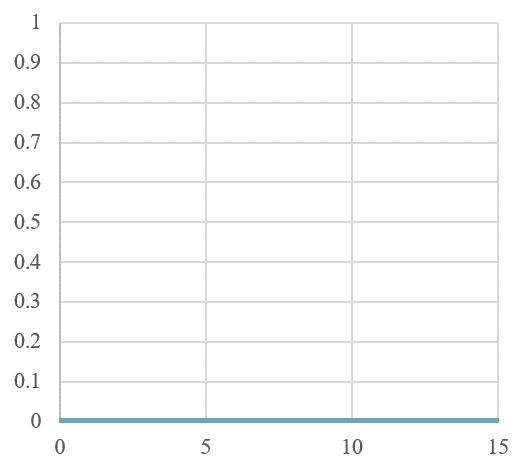}
			\label{fig:andg3sa}
		\end{minipage}%
	}
	\caption{Gradient amplitude $\left\| \frac{\partial L}{\partial s_l} \right\|$ of $l$-th block when $V_{th}=1, \alpha=3$ in the surrogate function $\sigma(x)$.}
	\label{fig:grad3}
\end{figure*}

\paragraph{Exploding Gradient}
Similar, here we set $V_{th}=1, \alpha=3$ in the surrogate function $\sigma(x)$ to analysis the exploding gradient, where  $\sigma'(1 - V_{th}) = 1.5 > 1$ and transmitting spikes to SNs is prone to causing exploding gradient. As shown in Figure~\ref{fig:grad3} (a-d), exploding gradient problem in DSNN with $\mathrm{OR}$, $\mathrm{XOR}$, $\mathrm{IAND}$, and $\mathrm{AND}$ is not serious.
Compared to Spiking ResNet without spike accumulation (Figure~\ref{fig:grad3} (d-h)), DSNN has a more gradual gradient.

\paragraph{Silence Problem of AND}
Moreover, some vanishing gradient happens in the Spiking ResNet with $\mathrm{AND}$ as shown in Figure~\ref{fig:andg2sa} and \ref{fig:andg3sa} (The gradient of $\mathrm{AND}$ is $0$), which is caused by the silence problem.
The attenuation of the gradient caused by the silent problem can be alleviated by the backpropagation of the spike accumulation path as shown in igure~\ref{fig:grad2} and \ref{fig:grad3}.
The gradients of DSNN with $\mathrm{OR}$, $\mathrm{XOR}$, $\mathrm{IAND}$, and $\mathrm{AND}$ increase slowly when propagating from deeper layers to shallower layers.

In general, DSNN could overcome the vanishing or exploding gradient problem well through spike accumulation.
At the same time, spike accumulation accumulates the output spikes of each block, thus increasing the discriminability of the features in network inference.

%% file: main.bbl
\begin{thebibliography}{63}
\providecommand{\natexlab}[1]{#1}
\providecommand{\url}[1]{\texttt{#1}}
\expandafter\ifx\csname urlstyle\endcsname\relax
  \providecommand{\doi}[1]{doi: #1}\else
  \providecommand{\doi}{doi: \begingroup \urlstyle{rm}\Url}\fi

\bibitem[Amir et~al.(2017)Amir, Taba, Berg, Melano, McKinstry, Di~Nolfo, Nayak,
  Andreopoulos, Garreau, Mendoza, Kusnitz, Debole, Esser, Delbruck, Flickner,
  and Modha]{amir2017low}
Amir, A., Taba, B., Berg, D., Melano, T., McKinstry, J., Di~Nolfo, C., Nayak,
  T., Andreopoulos, A., Garreau, G., Mendoza, M., Kusnitz, J., Debole, M.,
  Esser, S., Delbruck, T., Flickner, M., and Modha, D.
\newblock A low power, fully event-based gesture recognition system.
\newblock In \emph{Proceedings of the IEEE Conference on Computer Vision and
  Pattern Recognition (CVPR)}, pp.\  7243--7252, 2017.

\bibitem[Cao et~al.(2015)Cao, Chen, and Khosla]{cao2015spiking}
Cao, Y., Chen, Y., and Khosla, D.
\newblock Spiking deep convolutional neural networks for energy-efficient
  object recognition.
\newblock \emph{International Journal of Computer Vision}, 113\penalty0
  (1):\penalty0 54--66, 2015.

\bibitem[Chen \& Chen(2018)Chen and Chen]{chen2018saliency}
Chen, G. and Chen, Z.
\newblock Saliency detection by superpixel-based sparse representation.
\newblock In \emph{Advances in Multimedia Information Processing--PCM 2017:
  18th Pacific-Rim Conference on Multimedia, Harbin, China, September 28-29,
  2017, Revised Selected Papers, Part II 18}, pp.\  447--456. Springer, 2018.

\bibitem[Chen et~al.(2020)Chen, Qiao, Shi, Peng, Li, Huang, Pu, and
  Tian]{chen2020learning}
Chen, G., Qiao, L., Shi, Y., Peng, P., Li, J., Huang, T., Pu, S., and Tian, Y.
\newblock Learning open set network with discriminative reciprocal points.
\newblock In \emph{Computer Vision--ECCV 2020: 16th European Conference,
  Glasgow, UK, August 23--28, 2020, Proceedings, Part III 16}, pp.\  507--522.
  Springer, 2020.

\bibitem[Chen et~al.(2021{\natexlab{a}})Chen, Peng, Ma, Li, Du, and
  Tian]{chen2021amplitude}
Chen, G., Peng, P., Ma, L., Li, J., Du, L., and Tian, Y.
\newblock Amplitude-phase recombination: Rethinking robustness of convolutional
  neural networks in frequency domain.
\newblock In \emph{Proceedings of the IEEE/CVF International Conference on
  Computer Vision}, pp.\  458--467, 2021{\natexlab{a}}.

\bibitem[Chen et~al.(2021{\natexlab{b}})Chen, Peng, Wang, and
  Tian]{chenadversarial}
Chen, G., Peng, P., Wang, X., and Tian, Y.
\newblock Adversarial reciprocal points learning for open set recognition.
\newblock \emph{IEEE transactions on pattern analysis and machine
  intelligence}, 2021{\natexlab{b}}.
\newblock \doi{10.1109/TPAMI.2021.3106743}.

\bibitem[Comsa et~al.(2020)Comsa, Potempa, Versari, Fischbacher, Gesmundo, and
  Alakuijala]{comsa2020temporal}
Comsa, I.~M., Potempa, K., Versari, L., Fischbacher, T., Gesmundo, A., and
  Alakuijala, J.
\newblock Temporal coding in spiking neural networks with alpha synaptic
  function.
\newblock In \emph{International Conference on Acoustics, Speech and Signal
  Processing (ICASSP)}, pp.\  8529--8533. IEEE, 2020.

\bibitem[Deng et~al.(2009)Deng, Dong, Socher, Li, Li, and
  Fei-Fei]{deng2009imagenet}
Deng, J., Dong, W., Socher, R., Li, L.-J., Li, K., and Fei-Fei, L.
\newblock Imagenet: A large-scale hierarchical image database.
\newblock In \emph{2009 IEEE conference on computer vision and pattern
  recognition}, pp.\  248--255. Ieee, 2009.

\bibitem[Deng \& Gu(2021)Deng and Gu]{deng2021optimal}
Deng, S. and Gu, S.
\newblock Optimal conversion of conventional artificial neural networks to
  spiking neural networks.
\newblock In \emph{International Conference on Learning Representations
  (ICLR)}, 2021.
\newblock URL \url{https://openreview.net/forum?id=FZ1oTwcXchK}.

\bibitem[Deng et~al.(2022)Deng, Li, Zhang, and Gu]{deng2021temporal}
Deng, S., Li, Y., Zhang, S., and Gu, S.
\newblock Temporal efficient training of spiking neural network via gradient
  re-weighting.
\newblock In \emph{International Conference on Learning Representations}, 2022.

\bibitem[Ding et~al.(2019)Ding, Guo, Ding, and Han]{ding2019acnet}
Ding, X., Guo, Y., Ding, G., and Han, J.
\newblock Acnet: Strengthening the kernel skeletons for powerful cnn via
  asymmetric convolution blocks.
\newblock In \emph{Proceedings of the IEEE International Conference on Computer
  Vision}, pp.\  1911--1920, 2019.

\bibitem[Ding et~al.(2021)Ding, Zhang, Ma, Han, Ding, and Sun]{ding2021repvgg}
Ding, X., Zhang, X., Ma, N., Han, J., Ding, G., and Sun, J.
\newblock Repvgg: Making vgg-style convnets great again.
\newblock In \emph{Proceedings of the IEEE/CVF Conference on Computer Vision
  and Pattern Recognition}, pp.\  13733--13742, 2021.

\bibitem[Fang et~al.(2021{\natexlab{a}})Fang, Yu, Chen, Huang, Masquelier, and
  Tian]{fang2021deep}
Fang, W., Yu, Z., Chen, Y., Huang, T., Masquelier, T., and Tian, Y.
\newblock Deep residual learning in spiking neural networks.
\newblock \emph{Advances in Neural Information Processing Systems},
  34:\penalty0 21056--21069, 2021{\natexlab{a}}.

\bibitem[Fang et~al.(2021{\natexlab{b}})Fang, Yu, Chen, Masquelier, Huang, and
  Tian]{fang2020incorporating}
Fang, W., Yu, Z., Chen, Y., Masquelier, T., Huang, T., and Tian, Y.
\newblock Incorporating learnable membrane time constant to enhance learning of
  spiking neural networks.
\newblock In \emph{Proceedings of the IEEE/CVF International Conference on
  Computer Vision (ICCV)}, pp.\  2661--2671, 2021{\natexlab{b}}.

\bibitem[Fang et~al.(2022)Fang, Chen, Ding, Chen, Yu, Zhou, Tian, and other
  contributors]{SpikingJelly}
Fang, W., Chen, Y., Ding, J., Chen, D., Yu, Z., Zhou, H., Tian, Y., and other
  contributors.
\newblock Spikingjelly.
\newblock \url{https://github.com/fangwei123456/spikingjelly}, 2022.

\bibitem[Girshick et~al.(2014)Girshick, Donahue, Darrell, and
  Malik]{girshick2014rich}
Girshick, R., Donahue, J., Darrell, T., and Malik, J.
\newblock Rich feature hierarchies for accurate object detection and semantic
  segmentation.
\newblock In \emph{Proceedings of the IEEE Conference on Computer Vision and
  Pattern Recognition (CVPR)}, pp.\  580--587, 2014.

\bibitem[Han \& Roy(2020)Han and Roy]{han2020deep}
Han, B. and Roy, K.
\newblock Deep spiking neural network: Energy efficiency through time based
  coding.
\newblock In \emph{European Conference on Computer Vision (ECCV)}, pp.\
  388--404, 2020.

\bibitem[Han et~al.(2020)Han, Srinivasan, and Roy]{Han_2020_CVPR}
Han, B., Srinivasan, G., and Roy, K.
\newblock Rmp-snn: Residual membrane potential neuron for enabling deeper
  high-accuracy and low-latency spiking neural network.
\newblock In \emph{Proceedings of the IEEE/CVF Conference on Computer Vision
  and Pattern Recognition (CVPR)}, pp.\  13558--13567, 2020.

\bibitem[He et~al.(2016{\natexlab{a}})He, Zhang, Ren, and Sun]{he2015deep}
He, K., Zhang, X., Ren, S., and Sun, J.
\newblock Deep residual learning for image recognition.
\newblock In \emph{Proceedings of the IEEE/CVF Conference on Computer Vision
  and Pattern Recognition (CVPR)}, pp.\  770--778, 2016{\natexlab{a}}.

\bibitem[He et~al.(2016{\natexlab{b}})He, Zhang, Ren, and Sun]{he2016identity}
He, K., Zhang, X., Ren, S., and Sun, J.
\newblock Identity mappings in deep residual networks.
\newblock In \emph{European Conference on Computer Vision (ECCV)}, pp.\
  630--645. Springer, 2016{\natexlab{b}}.

\bibitem[Horowitz(2014)]{horowitz20141}
Horowitz, M.
\newblock 1.1 computing's energy problem (and what we can do about it).
\newblock In \emph{2014 IEEE International Solid-State Circuits Conference
  Digest of Technical Papers (ISSCC)}, pp.\  10--14. IEEE, 2014.

\bibitem[Hu et~al.(2018{\natexlab{a}})Hu, Tang, and Pan]{hu2018spiking}
Hu, Y., Tang, H., and Pan, G.
\newblock Spiking deep residual networks.
\newblock \emph{IEEE Transactions on Neural Networks and Learning Systems},
  2018{\natexlab{a}}.

\bibitem[Hu et~al.(2018{\natexlab{b}})Hu, Tang, Wang, and Pan]{hu2020spiking}
Hu, Y., Tang, H., Wang, Y., and Pan, G.
\newblock Spiking deep residual network.
\newblock \emph{arXiv preprint arXiv:1805.01352}, 2018{\natexlab{b}}.

\bibitem[Huh \& Sejnowski(2018)Huh and Sejnowski]{huh2017gradient}
Huh, D. and Sejnowski, T.~J.
\newblock Gradient descent for spiking neural networks.
\newblock In \emph{Advances in Neural Information Processing Systems
  (NeurIPS)}, pp.\  1440--1450, 2018.
\newblock URL
  \url{https://proceedings.neurips.cc/paper/2018/file/185e65bc40581880c4f2c82958de8cfe-Paper.pdf}.

\bibitem[Hunsberger \& Eliasmith(2015)Hunsberger and
  Eliasmith]{hunsberger2015spiking}
Hunsberger, E. and Eliasmith, C.
\newblock Spiking deep networks with lif neurons.
\newblock \emph{arXiv preprint arXiv:1510.08829}, 2015.

\bibitem[Hwang et~al.(2021)Hwang, Chang, Oh, Min, Jang, Park, Yu, Lee, and
  Park]{10.3389/fnins.2021.629000}
Hwang, S., Chang, J., Oh, M.-H., Min, K.~K., Jang, T., Park, K., Yu, J., Lee,
  J.-H., and Park, B.-G.
\newblock Low-latency spiking neural networks using pre-charged membrane
  potential and delayed evaluation.
\newblock \emph{Frontiers in Neuroscience}, 15:\penalty0 135, 2021.

\bibitem[Ioffe \& Szegedy(2015)Ioffe and Szegedy]{ioffe2015batch}
Ioffe, S. and Szegedy, C.
\newblock Batch normalization: Accelerating deep network training by reducing
  internal covariate shift.
\newblock In \emph{International conference on machine learning}, pp.\
  448--456. PMLR, 2015.

\bibitem[Kheradpisheh \& Masquelier(2020)Kheradpisheh and
  Masquelier]{kheradpisheh2020temporal}
Kheradpisheh, S.~R. and Masquelier, T.
\newblock Temporal backpropagation for spiking neural networks with one spike
  per neuron.
\newblock \emph{International Journal of Neural Systems}, 30\penalty0
  (06):\penalty0 2050027, 2020.

\bibitem[Kim et~al.(2018)Kim, Kim, Huh, Lee, and Choi]{KIM2018373}
Kim, J., Kim, H., Huh, S., Lee, J., and Choi, K.
\newblock Deep neural networks with weighted spikes.
\newblock \emph{Neurocomputing}, 311:\penalty0 373--386, 2018.

\bibitem[Kim et~al.(2020)Kim, Kim, and Kim]{kim2020unifying}
Kim, J., Kim, K., and Kim, J.-J.
\newblock Unifying activation- and timing-based learning rules for spiking
  neural networks.
\newblock In \emph{Advances in Neural Information Processing Systems
  (NeurIPS)}, pp.\  19534--19544, 2020.
\newblock URL
  \url{https://proceedings.neurips.cc/paper/2020/file/e2e5096d574976e8f115a8f1e0ffb52b-Paper.pdf}.

\bibitem[Krizhevsky et~al.(2009)Krizhevsky, Hinton,
  et~al.]{krizhevsky2009learning}
Krizhevsky, A., Hinton, G., et~al.
\newblock Learning multiple layers of features from tiny images.
\newblock 2009.

\bibitem[Krizhevsky et~al.(2012)Krizhevsky, Sutskever, and
  Hinton]{krizhevsky2012imagenet}
Krizhevsky, A., Sutskever, I., and Hinton, G.~E.
\newblock Imagenet classification with deep convolutional neural networks.
\newblock In \emph{Advances in Neural Information Processing Systems
  (NeurIPS)}, pp.\  1097--1105, 2012.
\newblock URL
  \url{https://proceedings.neurips.cc/paper/2012/file/c399862d3b9d6b76c8436e924a68c45b-Paper.pdf}.

\bibitem[Lee et~al.(2020)Lee, Sarwar, Panda, Srinivasan, and
  Roy]{lee2020enabling}
Lee, C., Sarwar, S.~S., Panda, P., Srinivasan, G., and Roy, K.
\newblock Enabling spike-based backpropagation for training deep neural network
  architectures.
\newblock \emph{Frontiers in Neuroscience}, 14, 2020.

\bibitem[Lee et~al.(2016)Lee, Delbruck, and Pfeiffer]{lee2016training}
Lee, J.~H., Delbruck, T., and Pfeiffer, M.
\newblock Training deep spiking neural networks using backpropagation.
\newblock \emph{Frontiers in Neuroscience}, 10:\penalty0 508, 2016.

\bibitem[Li et~al.(2017)Li, Liu, Ji, Li, and Shi]{cifar10_dvs}
Li, H., Liu, H., Ji, X., Li, G., and Shi, L.
\newblock Cifar10-dvs: An event-stream dataset for object classification.
\newblock \emph{Frontiers in Neuroscience}, 11:\penalty0 309, 2017.
\newblock ISSN 1662-453X.
\newblock \doi{10.3389/fnins.2017.00309}.
\newblock URL
  \url{https://www.frontiersin.org/article/10.3389/fnins.2017.00309}.

\bibitem[Li et~al.(2021)Li, Deng, Dong, Gong, and Gu]{pmlr-v139-li21d}
Li, Y., Deng, S., Dong, X., Gong, R., and Gu, S.
\newblock A free lunch from ann: Towards efficient, accurate spiking neural
  networks calibration.
\newblock In \emph{International Conference on Machine Learning (ICML)}, volume
  139, pp.\  6316--6325, 2021.
\newblock URL \url{https://proceedings.mlr.press/v139/li21d.html}.

\bibitem[Liu et~al.(2016)Liu, Anguelov, Erhan, Szegedy, Reed, Fu, and
  Berg]{liu2016ssd}
Liu, W., Anguelov, D., Erhan, D., Szegedy, C., Reed, S., Fu, C.-Y., and Berg,
  A.~C.
\newblock Ssd: Single shot multibox detector.
\newblock In \emph{European Conference on Computer Vision (ECCV)}, pp.\
  21--37. Springer, 2016.

\bibitem[Loshchilov \& Hutter(2017)Loshchilov and Hutter]{loshchilov2016sgdr}
Loshchilov, I. and Hutter, F.
\newblock {SGDR:} stochastic gradient descent with warm restarts.
\newblock In \emph{International Conference on Learning Representations
  (ICLR)}, 2017.
\newblock URL \url{https://openreview.net/forum?id=Skq89Scxx}.

\bibitem[Ma et~al.(2022)Ma, Peng, Chen, Zhao, Dong, and Tian]{ma2022picking}
Ma, L., Peng, P., Chen, G., Zhao, Y., Dong, S., and Tian, Y.
\newblock Picking up quantization steps for compressed image classification.
\newblock \emph{IEEE Transactions on Circuits and Systems for Video
  Technology}, 2022.

\bibitem[Maaten \& Hinton(2008)Maaten and Hinton]{maaten2008visualizing}
Maaten, L. v.~d. and Hinton, G.
\newblock Visualizing data using t-sne.
\newblock \emph{Journal of machine learning research}, 9\penalty0
  (Nov):\penalty0 2579--2605, 2008.

\bibitem[Meng et~al.(2022)Meng, Xiao, Yan, Wang, Lin, and
  Luo]{meng2022training}
Meng, Q., Xiao, M., Yan, S., Wang, Y., Lin, Z., and Luo, Z.-Q.
\newblock Training high-performance low-latency spiking neural networks by
  differentiation on spike representation.
\newblock In \emph{Proceedings of the IEEE/CVF Conference on Computer Vision
  and Pattern Recognition}, pp.\  12444--12453, 2022.

\bibitem[Mostafa(2017)]{mostafa2017supervised}
Mostafa, H.
\newblock Supervised learning based on temporal coding in spiking neural
  networks.
\newblock \emph{IEEE Transactions on Neural Networks and Learning Systems},
  29\penalty0 (7):\penalty0 3227--3235, 2017.

\bibitem[Neftci et~al.(2019)Neftci, Mostafa, and Zenke]{neftci2019surrogate}
Neftci, E.~O., Mostafa, H., and Zenke, F.
\newblock Surrogate gradient learning in spiking neural networks: Bringing the
  power of gradient-based optimization to spiking neural networks.
\newblock \emph{IEEE Signal Processing Magazine}, 36\penalty0 (6):\penalty0
  51--63, 2019.

\bibitem[Paszke et~al.(2019)Paszke, Gross, Massa, Lerer, Bradbury, Chanan,
  Killeen, Lin, Gimelshein, Antiga, Desmaison, Kopf, Yang, DeVito, Raison,
  Tejani, Chilamkurthy, Steiner, Fang, Bai, and Chintala]{PYTORCH}
Paszke, A., Gross, S., Massa, F., Lerer, A., Bradbury, J., Chanan, G., Killeen,
  T., Lin, Z., Gimelshein, N., Antiga, L., Desmaison, A., Kopf, A., Yang, E.,
  DeVito, Z., Raison, M., Tejani, A., Chilamkurthy, S., Steiner, B., Fang, L.,
  Bai, J., and Chintala, S.
\newblock Pytorch: An imperative style, high-performance deep learning library.
\newblock In \emph{Advances in Neural Information Processing Systems
  (NeurIPS)}, pp.\  8026--8037, 2019.
\newblock URL
  \url{https://proceedings.neurips.cc/paper/2019/file/bdbca288fee7f92f2bfa9f7012727740-Paper.pdf}.

\bibitem[Rathi \& Roy(2020)Rathi and Roy]{rathi2020dietsnn}
Rathi, N. and Roy, K.
\newblock Diet-snn: Direct input encoding with leakage and threshold
  optimization in deep spiking neural networks.
\newblock \emph{arXiv preprint arXiv:2008.03658}, 2020.

\bibitem[Rathi et~al.(2020)Rathi, Srinivasan, Panda, and
  Roy]{rathi2020enabling}
Rathi, N., Srinivasan, G., Panda, P., and Roy, K.
\newblock Enabling deep spiking neural networks with hybrid conversion and
  spike timing dependent backpropagation.
\newblock In \emph{International Conference on Learning Representations
  (ICLR)}, 2020.
\newblock URL \url{https://openreview.net/forum?id=B1xSperKvH}.

\bibitem[Redmon et~al.(2016)Redmon, Divvala, Girshick, and
  Farhadi]{redmon2016you}
Redmon, J., Divvala, S., Girshick, R., and Farhadi, A.
\newblock You only look once: Unified, real-time object detection.
\newblock In \emph{Proceedings of the IEEE Conference on Computer Vision and
  Pattern Recognition (CVPR)}, pp.\  779--788, 2016.

\bibitem[Roy et~al.(2019)Roy, Jaiswal, and Panda]{roy2019towards}
Roy, K., Jaiswal, A., and Panda, P.
\newblock Towards spike-based machine intelligence with neuromorphic computing.
\newblock \emph{Nature}, 575\penalty0 (7784):\penalty0 607--617, 2019.

\bibitem[Rueckauer et~al.(2017)Rueckauer, Lungu, Hu, Pfeiffer, and
  Liu]{Bodo2017Conversion}
Rueckauer, B., Lungu, I.-A., Hu, Y., Pfeiffer, M., and Liu, S.-C.
\newblock Conversion of continuous-valued deep networks to efficient
  event-driven networks for image classification.
\newblock \emph{Frontiers in Neuroscience}, 11:\penalty0 682, 2017.

\bibitem[Samadzadeh et~al.(2020)Samadzadeh, Far, Javadi, Nickabadi, and
  Chehreghani]{samadzadeh2021convolutional}
Samadzadeh, A., Far, F. S.~T., Javadi, A., Nickabadi, A., and Chehreghani,
  M.~H.
\newblock Convolutional spiking neural networks for spatio-temporal feature
  extraction.
\newblock \emph{arXiv preprint arXiv:2003.12346}, 2020.

\bibitem[Sengupta et~al.(2019)Sengupta, Ye, Wang, Liu, and
  Roy]{sengupta2019going}
Sengupta, A., Ye, Y., Wang, R., Liu, C., and Roy, K.
\newblock Going deeper in spiking neural networks: Vgg and residual
  architectures.
\newblock \emph{Frontiers in neuroscience}, 13:\penalty0 95, 2019.

\bibitem[Shrestha \& Orchard(2018)Shrestha and Orchard]{shrestha2018slayer}
Shrestha, S.~B. and Orchard, G.
\newblock Slayer: Spike layer error reassignment in time.
\newblock \emph{Advances in neural information processing systems}, 31, 2018.

\bibitem[Simonyan \& Zisserman(2015)Simonyan and Zisserman]{simonyan2015deep}
Simonyan, K. and Zisserman, A.
\newblock Very deep convolutional networks for large-scale image recognition.
\newblock In \emph{International Conference on Learning Representations
  (ICLR)}, 2015.
\newblock URL \url{http://arxiv.org/abs/1409.1556}.

\bibitem[St{\"o}ckl \& Maass(2021)St{\"o}ckl and Maass]{stockl2021optimized}
St{\"o}ckl, C. and Maass, W.
\newblock Optimized spiking neurons can classify images with high accuracy
  through temporal coding with two spikes.
\newblock \emph{Nature Machine Intelligence}, 3\penalty0 (3):\penalty0
  230--238, 2021.

\bibitem[Szegedy et~al.(2015)Szegedy, Liu, Jia, Sermanet, Reed, Anguelov,
  Erhan, Vanhoucke, and Rabinovich]{szegedy2015going}
Szegedy, C., Liu, W., Jia, Y., Sermanet, P., Reed, S., Anguelov, D., Erhan, D.,
  Vanhoucke, V., and Rabinovich, A.
\newblock Going deeper with convolutions.
\newblock In \emph{Proceedings of the IEEE/CVF Conference on Computer Vision
  and Pattern Recognition (CVPR)}, pp.\  1--9, 2015.
\newblock \doi{10.1109/CVPR.2015.7298594}.

\bibitem[Tavanaei et~al.(2019)Tavanaei, Ghodrati, Kheradpisheh, Masquelier, and
  Maida]{TAVANAEI201947}
Tavanaei, A., Ghodrati, M., Kheradpisheh, S.~R., Masquelier, T., and Maida, A.
\newblock Deep learning in spiking neural networks.
\newblock \emph{Neural Networks}, 111:\penalty0 47--63, 2019.

\bibitem[Wu et~al.(2018)Wu, Deng, Li, Zhu, and Shi]{wu2018STBP}
Wu, Y., Deng, L., Li, G., Zhu, J., and Shi, L.
\newblock Spatio-temporal backpropagation for training high-performance spiking
  neural networks.
\newblock \emph{Frontiers in Neuroscience}, 12:\penalty0 331, 2018.

\bibitem[Xiao et~al.(2022)Xiao, Meng, Zhang, He, and Lin]{xiaoonline}
Xiao, M., Meng, Q., Zhang, Z., He, D., and Lin, Z.
\newblock Online training through time for spiking neural networks.
\newblock In \emph{Advances in Neural Information Processing Systems}, 2022.

\bibitem[Xing et~al.(2019)Xing, Yuan, Huo, and
  Fang]{10.1007/978-3-030-36718-3_15}
Xing, F., Yuan, Y., Huo, H., and Fang, T.
\newblock Homeostasis-based cnn-to-snn conversion of inception and residual
  architectures.
\newblock In \emph{International Conference on Neural Information Processing},
  pp.\  173--184. Springer, 2019.

\bibitem[Zhang \& Li(2020)Zhang and Li]{zhang2020temporal}
Zhang, W. and Li, P.
\newblock Temporal spike sequence learning via backpropagation for deep spiking
  neural networks.
\newblock In \emph{Advances in Neural Information Processing Systems
  (NeurIPS)}, pp.\  12022--12033, 2020.
\newblock URL
  \url{https://proceedings.neurips.cc/paper/2020/file/8bdb5058376143fa358981954e7626b8-Paper.pdf}.

\bibitem[Zheng et~al.(2021)Zheng, Wu, Deng, Hu, and Li]{zheng2020going}
Zheng, H., Wu, Y., Deng, L., Hu, Y., and Li, G.
\newblock Going deeper with directly-trained larger spiking neural networks.
\newblock In \emph{Proceedings of the AAAI Conference on Artificial
  Intelligence}, volume~35, pp.\  11062--11070, 2021.
\newblock URL \url{https://ojs.aaai.org/index.php/AAAI/article/view/17320}.

\bibitem[Zhou et~al.(2021)Zhou, Li, Chen, Chandrasekaran, and
  Sanyal]{zhou2019temporal}
Zhou, S., Li, X., Chen, Y., Chandrasekaran, S.~T., and Sanyal, A.
\newblock Temporal-coded deep spiking neural network with easy training and
  robust performance.
\newblock In \emph{Proceedings of the AAAI Conference on Artificial
  Intelligence}, volume~35, pp.\  11143--11151, 2021.
\newblock URL \url{https://ojs.aaai.org/index.php/AAAI/article/view/17329}.

\bibitem[Zhou et~al.(2023)Zhou, Zhu, He, Wang, Yan, Tian, and
  Yuan]{zhou2023spikformer}
Zhou, Z., Zhu, Y., He, C., Wang, Y., Yan, S., Tian, Y., and Yuan, L.
\newblock Spikformer: When spiking neural network meets transformer.
\newblock 2023.

\end{thebibliography}
